\documentclass[aps,pre,superscriptaddress,groupedaddress]{revtex4-2} 
\usepackage{graphicx,color}  
\usepackage{bm}        
\usepackage{amssymb}   
\usepackage{bbold}
\usepackage{algorithm}
\usepackage{xcolor}
\usepackage{amsmath}

\usepackage{algpseudocode}
\newcommand{\MDGrevise}[1]{{\color{black}{#1}}}
\definecolor{ao(english)}{rgb}{0.0, 0.5, 0.0}

\newcommand{\ALrevise}[1]{\textcolor{black}{#1}}
\newcommand{\Koop}{\mathcal{K}}
\newcommand{\Kapp}{\mathbf{K}}

\newcommand{\reals}{\mathbb{R}}
\newcommand{\KDLAoo}{KDLA$_{\textrm{oo}}$}

\newcommand{\KDLoo}{KDL$_{\textrm{oo}}$}
\newcommand{\KDLso}{KDL$_{\textrm{so}}$}

\usepackage[normalem]{ulem}

\begin{document}

 \title{
Enhancing Predictive Capabilities in Data-Driven Dynamical Modeling with Automatic Differentiation: Koopman and Neural ODE Approaches 
}

\author{C. Ricardo Constante-Amores$^1$, Alec J. Linot$^2$, Michael D. Graham$^1$}\email{mdgraham@wisc.edu}
\affiliation{$^1$Department of Chemical and Biological Engineering, University of Wisconsin-Madison, Madison WI 53706, USA\\
$^2$Department of Mechanical and Aerospace Engineering, University of California, Los Angeles, CA 90095, USA}
% \author{Alec J. Linot}

% \author{Michael D. Graham}
% \email{mdgraham@wisc.edu}
% \affiliation{Department of Chemical and Biological Engineering, University of Wisconsin-Madison, Madison WI 53706, USA}
% \email{mdgraham@wisc.edu}
\date{\today}

\begin{abstract}
Data-driven approximations of the Koopman operator are promising for predicting the time evolution of systems characterized by complex dynamics. Among these methods, the approach known as extended dynamic mode decomposition with dictionary learning (EDMD-DL) has garnered significant attention.  Here we present a modification of EDMD-DL that concurrently determines both the dictionary of observables and the corresponding approximation of the Koopman operator. This innovation leverages automatic differentiation to facilitate gradient descent computations through the pseudoinverse. We also address the performance of several alternative methodologies. We assess a 'pure' Koopman approach, which involves the direct time-integration of a linear, high-dimensional system governing the dynamics within the space of observables. Additionally, we explore a modified approach where the system alternates between spaces of states and observables at each time step -- this approach no longer satisfies the linearity of the true Koopman operator representation. For further comparisons, we also apply a state space approach (neural ODEs). We consider systems encompassing two and three-dimensional ordinary differential equation systems featuring steady, oscillatory, and chaotic attractors, as well as partial differential equations exhibiting increasingly complex and intricate behaviors. Our framework significantly outperforms EDMD-DL. Furthermore, the state space approach offers superior performance compared to the 'pure' Koopman approach where the entire time evolution occurs in the space of observables.  When the temporal evolution of the Koopman approach alternates between states and observables at each time step, however, its predictions become comparable to those of the state space approach.

\end{abstract}

\maketitle

\begin{quote}
\textbf{Significance statement.} For every dynamical system, there is a linear
% , but  infinite-dimensional
Koopman operator that describes the evolution of an arbitrary observable. This concept has garnered significant attention within the scientific community, driven by the wealth of available data and its profound connection to dynamic mode decomposition (DMD). This work enhances the  Extended DMD with Dictionary Learning method by using automatic differentiation to simultaneously find both the Koopman operator and dictionary of observables for a diverse range of systems. These systems encompass two and three-dimensional ordinary differential equation systems featuring steady, oscillatory, and chaotic attractors, as well as partial differential equations exhibiting increasingly complex and intricate behaviors. We also explore variants of this approach that move between spaces of states and observables, and compare the performance of these approaches relative to a state space approach using  neural ODEs.
\end{quote}

\section{Introduction }

Dynamical models are a valuable tool for understanding and solving problems in various fields \cite{strogatz}. They allow us to mathematically represent the behavior of complex systems and study their dynamics over time, shedding light on how they will behave under different conditions \citep{brin2002introduction}.   
The construction of data-driven dynamical models has drawn significant interest in the literature since the  
increase in data availability from high-fidelity simulations or high-speed imaging; 
\MDGrevise{ see comprehensive reviews by e.g., \citet{Hatfield,Karniadakis,mezic_2013}}. However, data-driven modeling remains challenging
when dealing with nonlinear, high-dimensional, and complex dynamics. The nonlinearity and high dimensionality of real-world phenomena can make it difficult to identify the underlying relationships and patterns in the data. Additionally, a lack of prior knowledge of the physical laws of the system can make it difficult to develop accurate models for forecasting. To overcome these challenges, machine learning techniques, such as deep learning, are often used in combination with large amounts of data and computational resources  \citep{Pathak,alec_coutte,Lee}.

\textcolor{black}{A majority of the methods applied to dynamical systems are based on the {\it state space approach} \citep{strogatz}}.
However, almost a century ago, Koopman proposed an alternative approach which we refer to as the \textcolor{black}{{\it function space approach}
\citep{koopman1931hamiltonian,Budisic,mezic_2013}. The function space approach } 
involves investigating the evolution of functions (or observables) of the state space, and analyzing the decomposition into a basis of eigenfunctions of the Koopman operator. This approach is useful as it allows us to describe \emph{nonlinear} dynamics
\textcolor{black}{in an infinite-dimensional space of observable functions  via a  }
% with an infinite-dimensional
\emph{linear} Koopman operator that evolves observables forward in time. 
% \footnote{\textcolor{black}{In this work, we will refer to the Koopman operator as infinite-dimensional for convenience. However, it is important to acknowledge that while the Koopman operator is often linked to infinite-dimensional function spaces, it does not inherently possess a dimension itself. A more precise statement would be that the Koopman operator is commonly defined to act on infinite-dimensional spaces. 
As such, the spectral information of the Koopman operator can provide valuable insights into the dynamics of a system \citep{Mezi2005}.
The linearization of a nonlinear system brings the downside that the 
Koopman operator \textcolor{black}{acts in an }
infinite-dimensional \textcolor{black}{ space}; thus there is a trade-off between linearity and dimensionality. \textcolor{black}{ 
Koopman methods offer distinct advantages over state-space approaches, primarily attributed to the Koopman operator's ability to offer a linear representation of global dynamics within inherently nonlinear dynamical systems. By decomposing the system's evolution into linear dynamics, the Koopman operator simplifies the understanding of the system's behavior through the analysis of eigenfunctions and eigenvalues \citep{Duraisamy,mezic_2013}. Moreover, the linear representation facilitates control using well-established techniques, such as Linear Quadratic Regulator (LQR), resulting in efficacious optimal control for nonlinear systems \citep{BEVANDA202213,
mamakoukas2019local}. }

The notion of converting nonlinear dynamical systems into linear dynamical systems has fascinated the scientific community; but it was not until recent decades, with the emergence of data-driven research techniques, such as Dynamic Mode Decomposition (DMD) that this function space approach gained widespread popularity \cite{Lusch,Qianxiao,brunton2016koopman,kDMD,Wilson,mpEDMD}. DMD  is based on the concept of finding a linear operator that best maps (in a least-squares sense)  equispaced snapshots of data. Despite the increased attention given to the function space approach, most efforts have been restricted to low-dimensional systems, as the computational complexity of analysing high-dimensional systems increases dramatically with the number of state variables \cite{eDMD,dmd_book,Alford,fox2023predicting}.
The linear transformation of a system is done through a set of observables that are a function of the system's state space. Once, the observables are identified, the Koopman operator can be approximated whose spectral analysis resulting in 
Koopman eigenfunctions, eigenvalues, and modes.
\textcolor{black}{The Koopman eigenfunctions define a set of natural coordinates. In the eigenfunction basis, the dynamics along different eigenfunction directions are decoupled,
making it easier to analyze and understand the system's behavior in these new coordinate directions.} The Koopman eigenvalues dictate the temporal dynamics, while the Koopman modes enable the reconstruction of the initial state space from the linear representation of the observables.

Various methods have been proposed to approximate the Koopman operator, but the leading method is the DMD algorithm introduced by \citet{schmid_2010} and connected to the Koopman operator by \citet{rowley_2009}. 
\textcolor{black}{ The original DMD  considered the state as the only observable, making it essentially a linear state space model. To establish a direct link to Koopman theory, one must move beyond traditional DMD and consider the lifting of the state into a higher-dimensional space. Thus, instead of solely considering the state as the set of observables, a broader set of observables, which can be nonlinear functions of the state, is incorporated. }
Various strategies have been proposed to ensure that the input function space is sufficiently ‘rich’ (e.g., ‘kernel’-based methods, see \citet{Kutz_2016,colbrook_ayton_2023}). 
Alongside DMD, several data-driven methodologies have been proposed to approximate the Koopman operator and identify its eigenvalues. Those methodologies include the Extended-DMD (EDMD) and the kernel-DMD (kDMD), which make use of a fixed dictionary and kernel functions, respectively \citep{eDMD,kDMD}. 
\citet{Kutz_2016} showed that choosing an appropriate set of observables is crucial to finding an accurate approximation of the Koopman operator. 
This statement is particularly true for high-dimensional systems, where constructing and evaluating an explicit dictionary of observables can be prohibitively expensive \citep{lrans}.

\textcolor{black}{ While the study of dynamical systems using the 
% infinite-dimensional 
Koopman operator holds significant promise, its approximation results in 
substantial challenges. The main difficulty arises primarily because it acts on infinite-dimensional spaces, and the truncation for its approximation results in the loss of its continuous spectrum (e.g., which results in a lack of uncountable eigenvalues and eigenfunctions) \citep{Redei,Mezi2005,mpEDMD,resdmd}. Continuous spectra are characteristic ofchaotic systems such as turbulent flows. Various methods have been developed to maintain the continuous nature of the spectrum, such as the `residual DMD' (ResDMD) 
% and the `measure preserving EDMD' (mpEDMD) 
standing out as the most effective approach \citep{resdmd}. The ResDMD framework generates smoothed approximations of spectral measures, including continuous spectra.}
% mpEDMD, the framework is based onorthogonal Procrustes problem that enforces measure-preserving approximations of Koopman operator \citep{procrustes}.}

Recently, there have been interesting attempts to use deep neural networks to approximate the 
% infinite-dimensional 
Koopman operator 
(\citet{Lusch,lrans,Qianxiao}). 
% These attempts have so far been restricted to low-dimensional examples. 
In \MDGrevise{a method they denoted} ``deepKoopman", \citet{Lusch}  used an encoder/decoder to identify  linear embeddings for relatively low-dimensional
nonlinear systems \textcolor{black}{(e.g., two and three-dimensional ordinary differential equations)}. This method requires an auxiliary neural network to be coupled with the latent space to parameterize the continuous eigenvalue spectrum of the Koopman operator \textcolor{black}{as they focus on cases with conserved quantities (e.g., Hamiltonian systems)}. 
In 
`LARNS' \textcolor{black}{(Linearly Recurrent Autoencoder Networks)}, \citet{lrans}
\textcolor{black}{ combined a neural network architecture that is essentially an autoencoder with linear recurrence}.
\citet{koopman_thesis} and \citet{lrans} have shown \MDGrevise{some} success in predicting the full state \MDGrevise{dynamics} of high-dimensional systems such as the Kuramoto-Sivashinsky Equation \textcolor{black}{ (KSE)} with beating traveling waves.
\ALrevise{Finally, the approach of particular interest here, is EDMD with dictionary learning (EDMD-DL). In EDMD-DL neural networks are trained to map the state to a set of observables, which are evolved forward with the Koopman operator. The dictionary of observables and the Koopman operator are learned together in this method through }
an alternating optimization scheme that consists of sequential execution of two steps: \textcolor{black}{approximating the Koopman operator } 
for a fixed dictionary of observables, then updating the dictionary with the fixed approximate Koopman operator. \ALrevise{Unfortunately, this alternating approach can lead to suboptimal solutions as we elaborate on in Section \ref{duffing_section}.}

In this study, we propose a variant of the EDMD-DL approach that significantly improves  predictive capabilities by avoiding this alternating optimization procedure. The key idea of our approach is to replace the approximation of the Koopman operator with the best linear fit using the Moore-Penrose pseudo-inverse, and then perform automatic differentiation through this pseudo-inverse to compute the gradient used in updating the neural networks (i.e.\ dictionary). 
We show that our approach 
\MDGrevise{outperforms} other Koopman-based approaches, \textcolor{black}{ and we also explore a modified time-evolution approach where the system alternates between the spaces of states and observables.}
Additionally, we compare 
the function space approach with a state space approach, where we train nonlinear neural ODEs \ALrevise{to evolve the state forward in time.}
The rest of this article is organized as follows: Section 2 presents the framework of our methodology. Section 3 provides \ALrevise{results for many examples of increasing complexity.}
Finally, concluding remarks are summarised in Section 4.

\section{Problem statement and mathematical background}

\begin{figure}
\centering
\includegraphics[width=1.0\textwidth]{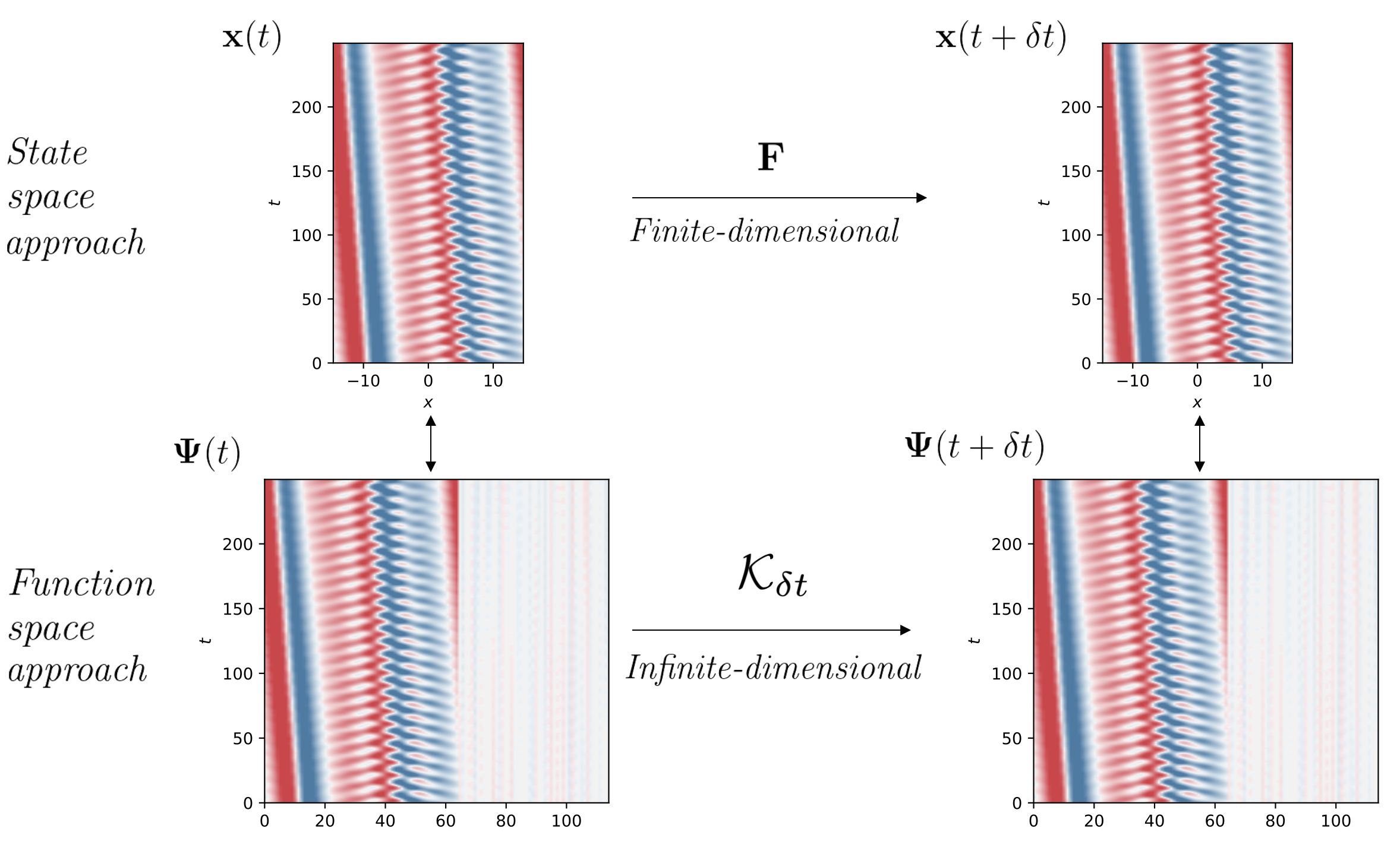}
\caption{\label{intro}Schematic representation of the \textcolor{black}{ {\it state space} and {\it function space approaches}} for the  Kuramoto Sivashinsky Equation with a beating traveling wave. The top path evolves the state space $\mathbf{x }$  using a finite-dimensional operator,  $\mathbf{F }(t +\delta t) $, while the bottom path updates the observables  $\mathbf{\Psi }$ via the
% an infinite dimension 
Koopman operator, $\mathbf{K }(t +\delta t)$. 
The vector of observables $\mathbf{\Psi }$ is a function of the full state and a set of dictionary elements from a deep neural network.
\textcolor{black}{For this particular case, the PDE has been solved in a grid of 64 points, and 50 trainable dictionary elements complete the set of observables.}}
\end{figure}

In this section, we introduce the framework for the  state and function space approaches used in this study. Figure \ref{intro} serves as a visual summary of this section, illustrating key concepts for the  Kuramoto-Sivashinsky with beating traveling waves. In the top row, we depict the direct evolution of states, while the bottom row illustrates the evolution of observables by the Koopman operator.

\subsection{State space approach }

We consider systems that display deterministic, Markovian dynamics,
\ALrevise{which may either be represented by a discrete-time map}
\begin{equation}
    {\bf x}(t+ \delta t )=\mathbf{F}({\bf x} (t))
\end{equation}
or an ordinary differential equation (ODE)
\begin{equation}
{\color{black}{\dot{\textbf{x}}=\textbf{f}(\textbf{x})}},
\label{cont-represent}
\end{equation}
\ALrevise{where $\textbf{x}(t) \in \reals^n$ is the state of the system.}
The aim \ALrevise{of data-driven modeling} is to find an accurate representation of $\mathbf{F}$ or $\textbf{f}$. 
\ALrevise{Although other works have shown neural networks can approximate $\mathbf{F}$ \citep{alec_pre}, here we choose to approximate the continuous-time representation $\bf{f}(\textbf{x})$.}
We use the framework presented by \citet{chen2019neural}, known as ``neural ODEs''\ALrevise{, in which $\bf{f}$ is approximated as a neural network with weights $\theta_f$}.
\ALrevise{By integrating this neural network representation of  $\bf{f}$ forward in time from $t_i$ to $t_i+\delta t$ we can approximate the state at this time with}
\begin{equation}\label{}
	\tilde{{\bf x}}(t_i+\delta t)={\bf x}(t_i)+\int_{t_i}^{t_i+\delta t}{\bf f}({\bf x};\theta_f) dt,
\end{equation}
\textcolor{black}{here $\tilde{{\bf x}}((t_i+\delta t)$ represent the state variable predictions}.
\ALrevise{Using this prediction, we then train these neural ODEs }
to minimize the difference between the prediction of the state and the known data
\begin{equation} \label{}
	\mathcal{L}=\frac{1}{nN}\sum_{i=1}^N||{\bf x}(t_i+\delta t)-\tilde{{\bf x}}(t_i+\delta t)||_2^2.
\end{equation}
The derivatives of $\bf{f}$ with respect to the parameters can be determined either by solving an adjoint problem or using automatic differentiation \citep{Asch}.
In \citet{alec_chaos,alec_coutte}, the authors applied this methodology to predict the chaotic dynamics in the Kuramoto-Sivashinski equation and turbulent minimal Couette flow. They showed that neural ODEs \ALrevise{track trajectories of these systems over short times and match statistics of these systems at long times.}

\subsection{Function space approach }

\ALrevise{Next, we consider an alternative view of time evolution in terms of a set of observables $\mathbf{\Psi }({\bf x})$. These observables can be evolved forward in time by}

\begin{equation}
   \mathbf{\Psi }( {\bf x}(t + \delta t))=\Koop_{\delta t}(\mathbf{\Psi }( {\bf x}(t))=\mathbf{\Psi }(\mathbf{F}({\bf x}(t))),
\end{equation}
where the mapping $\Koop_{\delta t}$ is the 
% infinite-dimensional 
{\it Koopman operator}. It is an operator on {\it functions}
\textcolor{black}{that applies to \emph{all} observables, of which there are infinitely many.}
To relate $\Koop_{\delta t}$ with the original differential equation, and understand how  $\mathbf{\Psi }$ evolves with time, we consider
\begin{equation}
\mathbf{\Psi } ( {\bf x}(t + \delta t)) - \mathbf{\Psi } ( {\bf x}(t)),
\end{equation}
which represents  the change in $\mathbf{\Psi }$ over the time interval $\delta t$;  when $\delta t \rightarrow 0$, the following statement holds
\begin{equation}
    \lim_{\delta t \rightarrow 0}\frac{\mathbf{\Psi } ( {\bf x}(t + \delta t)) - \mathbf{\Psi } ( {\bf x}(t))}{\delta t}=\frac{\partial  \mathbf{\Psi }}{ \partial  t}=\frac{\mathbf{\Psi }({\textbf{x}}(t)) +
    \dot{\textbf{x}} \cdot \frac{\partial}{\partial \textbf{x}}\mathbf{\Psi } \delta t +
    O(\delta t ^2)-  \mathbf{\Psi }(\textbf{x}(t))}{\delta t}=\dot{\textbf{x}} \cdot \frac{\partial}{\partial \textbf{x}}\mathbf{\Psi } =
    \textbf{f} \cdot \frac{\partial}{\partial \textbf{x}}\mathbf{\Psi },
\end{equation}
so 
\textcolor{black}{
\begin{equation}
    \frac{\partial  \mathbf{\Psi }}{ \partial  t} =
    \textbf{f} \cdot \frac{\partial}{\partial \textbf{x}}\mathbf{\Psi }.
\end{equation}}
Solving this with initial condition $\mathbf{\Psi }({\bf x}(t ))$ yields  $\mathbf{\Psi }({\bf x}(t + \delta t))$ (i.e., integrating this gives the Koopman operator). If ${\bf x}(t)$ is a trajectory on an attractor, than any observable $\mathbf{\Psi }( {\bf x}(t))$ will neither decay nor grow as $t \rightarrow \infty$, since
\begin{equation*}
\mathbf{\Psi }({\bf x}(t + \delta t))=\Koop_{\delta t}(\mathbf{\Psi }( {\bf x}(t))).
\end{equation*}
Thus the time series of $\mathbf{\Psi }$ will be a superposition of the eigenfunctions of $\Koop_{\delta t}$, whose eigenvalues $\lambda_k$ have unit magnitude, i.e. $|\lambda_k |=1$. Any contributions from an eigenfunction with  $|\lambda_k |<1$ decay as $t \rightarrow \infty$, and 
\textcolor{black}{the fact that the dynamics live on an attractor},
prohibits any $\lambda_k$  from having 
$|\lambda_k |\geq1$.
Unlike \MDGrevise{with principal components analysis (PCA)},
there is no significance to the values of $\lambda$ in terms of their contribution to the signal (e.g., there is no ordering that a priori gives relative energy content).

Therefore, the nonlinear dynamical operator $\textbf{f}$, which is finite-dimensional, and the Koopman operator $\Koop_{\delta t}$, which acts in infinite-dimensional \textcolor{black}{spaces}, both capture the same underlying dynamics as they are two different but equivalent representations of the same dynamics.

\subsubsection{Finite approximation of the Koopman operator \label{koopman_approx}}

% \textcolor{black}{ The Koopman operator is 
% % an infinite-dimensional
% linear operator that captures the essential aspects  of a nonlinear dynamical system. However, the Koopman operator does not capture every detail of a nonlinear system, as  explored in the work of \citet{Redei}, who demonstrated spectral isomorphism does not imply spacial isomorphism.}
%
To approximate the Koopman operator, 
we only have access to equispaced time-series data (i.e., we only have information of ${\bf x}(t)$ and ${\bf x}(t + \delta t)$ at discrete time intervals).  The 
% infinite-dimensional
Koopman operator  $\Koop_{\delta t}$ will be approximated by a matrix-valued Koopman operator  $\Kapp$.
\textcolor{black}{This approximation seeks to maintain the essential properties of the 
% infinite-dimensional 
operator while making it computationally feasible. It is important to note that while this preservation is desirable, it is not an inevitable outcome.}
Finally, we need to carefully select a set of observables $(\mathbf{\Psi }( {\bf x}(t))$ that can accurately capture the nonlinear dynamics of  $\mathbf{f}$.

The EDMD  framework requires a data set of $M$ snapshot pairs, and a dictionary of observables. So, we can construct a matrix of observables, whose columns are the vector of observables \ALrevise{(snapshots)} at different times 
\begin{equation}
\boldsymbol{\psi}(t)=\left [ \boldsymbol{\Psi}(\textbf{x}(t_1))~~ \boldsymbol{\Psi}(\textbf{x}(t_2)) ~~\ldots \right ],
\end{equation}
and a corresponding matrix with the observables now evaluated $t + \delta t$ later,
\begin{equation}
\boldsymbol{\psi}({t+\delta t}) =\left [ \boldsymbol{\Psi}(\textbf{x}(t_1 + \delta t))~~ \boldsymbol{\Psi} (\textbf{x} (t_2 + \delta t)) ~~\ldots  \right ].
\end{equation}
\MDGrevise{The original EDMD methodology took the observables to be Hermite or Legendre polynomial functions of the state.}
Finding an approximate (matrix) Koopman operator $\Kapp$ from the matrix of observables, with fixed dictionary elements, can be done by solving the following optimisation problem
\begin{equation} 
   \underset{\Kapp}
   {\text{min}} ||\boldsymbol{\psi}{(t + \delta t )} -\Kapp \boldsymbol{\psi}{(t )}||_F^2,  \label{EDMD_original}
\end{equation}
where $|| \cdot||_F$ refers to the Frobenius norm. The solution of this minimisation problem is \ALrevise{given by}
\begin{equation} \label{eq:K}
\Kapp= \boldsymbol{G}^+ \boldsymbol{A}  
\end{equation}
\noindent
where $+$ superscript denotes the pseudo (Moore-Penrose) inverse and
\begin{equation}
\begin{aligned}
\boldsymbol{G} &= \frac{1}{M} \sum_{m=1}^M \boldsymbol{\psi}(t)^* \boldsymbol{\psi}(t), \\
\boldsymbol{A} &= \frac{1}{M} \sum_{m=1}^M \boldsymbol{\psi}(t)^* \boldsymbol{\psi}(t + \delta t),
\end{aligned}
\end{equation}

\ALrevise{As mentioned in the Introduction, }
\textcolor{black}{\citet{Qianxiao} proposed the EDMD with dictionary learning (EDMD-DL) \ALrevise{method}, in which they allowed the dictionary to be trainable via deep neural networks (i.e. $\mathbf{\Psi }({\bf x }(t))$ becomes $\mathbf{\Psi }({\bf x }(t);\theta)$, where $\theta$ represents the weights of the neural network).} The EDMD-DL algorithm involves a mapping to the observable space $\boldsymbol{\Psi}=\boldsymbol{\chi}(\mathbf{x})$, and a mapping back to the full state $\mathbf{x}=\tilde{\boldsymbol{\chi}}(\boldsymbol{\Psi})$ \ALrevise{along with finding the linear map $\Kapp$}. 
\ALrevise{Unlike standard EDMD, this changes the optimization problem to
\begin{equation}
   \underset{\Kapp, \boldsymbol{\psi}}
   {\text{min}} ||\boldsymbol{\psi}{(t + \delta t )} -\Kapp \boldsymbol{\psi}{(t )}||_F^2. \label{EDMD-DL} 
\end{equation} In the EDMD-DL algorithm this minimum is sought through an iterative process in which the dictionary $\boldsymbol{\Psi}$ is fixed and $\Kapp$ is found using the pseudo-inverse, then $\Kapp$ is fixed and standard neural network training is performed.} 
These two steps are iterated until convergence. 
To improve the stability of the algorithm, a Tikhonov regularization is  introduced in the minimisation.
\MDGrevise{For brevity, we will call this method `KDL', and introduce specific implementations and variants of it below.}

\ALrevise{Here we improve upon this algorithm by avoiding this iterative procedure entirely. Instead, we optimize the NN parameters by using automatic differentiation to compute the gradient through the pseudoinverse directly. We do this by combining Eq.\ \ref{eq:K} and Eq.\ \ref{EDMD-DL} resulting in 
\begin{equation} \label{eq:min}
   \underset{\boldsymbol{\psi}}
   {\text{min}} ||\boldsymbol{\psi}{(t + \delta t )} -\boldsymbol{G}^+ \boldsymbol{A} \boldsymbol{\psi}{(t )}||_F^2.  
\end{equation}
By directly seeking the optimum of the neural network weights, and noting that $\Kapp$ already comes from the pseudoinverse, we never have to perform the iterative procedure in \citep{Qianxiao}.}

Next, we explain the loss function, $\mathcal{L}$, used in our approach when training the neural networks to find the dictionary and $\Kapp$. We aim to minimize 
\begin{equation}
\mathcal{L}({\bf x },\theta)=\left \|  \boldsymbol{\psi} (\mathbf{x},t+\delta t; \theta) - \tilde{\boldsymbol{\psi}} (\mathbf{x},t+\delta t; \theta)   \right \|_F,   
\end{equation}
where $ \tilde{\boldsymbol{\psi}} (t+\delta t) $ represents the observables predictions. Here,
$ \tilde{\boldsymbol{\psi}} (t+\delta t) = \Kapp \boldsymbol{\psi} (t) $, and $\Kapp= \boldsymbol{\psi}(t+\delta t)\boldsymbol{\psi}(t)^+$ \ALrevise{upon simplifying Eq.\ \ref{eq:K}}. In scenarios where stochastic gradient descent \ALrevise{methods are} employed with a batch size that does not align with the dataset size, we introduce two distinct sets of \ALrevise{observables}. We define $\boldsymbol{\psi}_{\text{NN}}$ as the \ALrevise{observables} associated with the linear time evolution, and $\boldsymbol{\psi}_{K}$ as \ALrevise{observables used to construct $\Kapp$.} 
Under these conditions, the loss takes on the following form:
\begin{equation}
\mathcal{L}({\bf x },\theta)=\left \|   \boldsymbol{\psi} (\mathbf{x}, t+\delta t; \theta)  -  \tilde{\boldsymbol{\psi}} (\mathbf{x}, t+\delta t; \theta)    \right \|_F= 
 \left \|   \boldsymbol{\psi}_{\text{NN}} (\mathbf{x}, t+\delta t; \theta)  -  \boldsymbol{\psi}_K (\mathbf{x}, t+\delta t; \theta) \boldsymbol{\psi}_K(\mathbf{x}, t; \theta)^+  \boldsymbol{\psi}_{\text{NN}} (\mathbf{x}, t; \theta)  \right \|_F.
 \label{loss_batch}
\end{equation}
\ALrevise{In some scenarios, this distinction may be important. By selecting a large number of data points to construct $\boldsymbol{\psi}_{K}$ the linear map $\Kapp$ would vary less during training, and the advantages of stochasticity could be introduced through a small batch size in the selection of $\boldsymbol{\psi}_{\text{NN}}$. However, for the remainder of this work we choose a batch size equivalent to the data size (i.e. $\boldsymbol{\psi}_K=\boldsymbol{\psi}_{\text{NN}}$).} 

\ALrevise{When these two sets of observables are of the same size we may simplify the loss in Eq.\ \ref{loss_batch} to}
\begin{equation} \label{loss}
\mathcal{L}({\bf x },\theta) = ||  \boldsymbol{\psi}{({\bf x }, t + \delta t;\theta )(\mathcal{I}- (\boldsymbol{\psi}({\bf x },t;\theta)^+  \boldsymbol{\psi}({\bf x }, t;\theta))} ||_F,
\end{equation}
\ALrevise{where $\mathcal{I}$ is the identity.}
It is important to note that using this formulation the number of snapshots in $\boldsymbol{\psi}({\bf x }, t;\theta))$ \emph{must} exceed the number of observables. If the number of \textcolor{black}{observables} exceeds the snapshots  $\boldsymbol{\psi}({\bf x },t;\theta)^+  \boldsymbol{\psi}({\bf x }, t;\theta)=\mathcal{I}$ and the loss simply evaluates to 0. This highlights another reason why the formulation of the loss in Eq.\ \ref{loss_batch} may be preferable in some scenarios.

Automatic differentiation enables us to compute the gradient of Eq. \ref{loss} 
with respect to the neural network weights for $\boldsymbol{\psi}$.
% The advance that was required was the availability of a differentiable function for SVD, and consequently, the pseudoinverse. 
We use PyTorch for the pseudoinverse calculation, which uses the approach described in  \citet{golub1973differentiation}  and \citet{Decell}. Both works showed that the derivative of the pseudoinverse can be computed without the need to ever compute the gradient through the SVD. Instead, the derivative of the pseudoinverse can be computed using this result:
\begin{equation}
dA^+=-A^+dAA^++(\mathcal{I}-A^+A)dA^*A^{+*}A^++A^+A^{+*}dA^*(\mathcal{I}-AA^+).
\end{equation}
This means we only need the gradient of the matrix of observables with respect to our network parameters, and the pseudoinverse of the observables to compute the gradient of the pseudoinverse with respect to the network parameters, such as
\begin{equation}
\frac{\partial\boldsymbol{\psi}^+}{\partial\theta}=-\boldsymbol{\psi}^+
\frac{\partial\boldsymbol{\psi}}{\partial\theta}\boldsymbol{\psi}^+
+(\mathcal{I}-\boldsymbol{\psi}^+\boldsymbol{\psi})
\frac{\partial\boldsymbol{\psi}^*}{\partial\theta} \boldsymbol{\psi}^{+*}\boldsymbol{\psi}^++\boldsymbol{\psi}^+\boldsymbol{\psi}^{+*}\frac{\partial\boldsymbol{\psi}^*}{\partial\theta}(\mathcal{I}-\boldsymbol{\psi}\boldsymbol{\psi}^+).
\end{equation}
For our case,  the term $(\mathcal{I}-\boldsymbol{\psi}\boldsymbol{\psi}^+)$ equals 0, although PyTorch still computes it. Finally, the computational difficulties associated with computing gradients of the SVD are not an issue in this computation. The only areas where we could face numerical difficulties are in either the computation of the gradient of the matrix or in the computation of the pseudoinverse. The gradient computation forms the backbone of all neural network training and the pseudoinverse comes from computing the SVD, so both of these are widely used and there is not an obvious place where we might expect the algorithm to be unstable.

% We use PyTorch to perform the pseudoinverse calculation \citep{paszke2019pytorch}. Many works have shown how to compute the gradient of the pseudoinverse, and the numerical issues associated with this procedure \citep{Papadopoulo,ionescu2015matrix}. When computing the gradient of the SVD the procedure becomes numerically unstable when there are repeated singular values \citep{Papadopoulo}. For the cases we consider, we did not observe any issues during the gradient calculation (see Appendix A for the singular values for the Koopman operator trained on the Duffing oscillator). We note that different methods have been proposed to avoid gradient problems such as SVD-TopN (that keeps only the large top-N singular values) or SVD-Taylor which uses a Taylor polynomial to decompose the gradient function \citep{SVD_gradi}.}

Table \ref{Table_NN} summarizes 
the architecture and parameters utilized in the studies of this paper for \KDLAoo.
We train each neural network until the loss levels off using an Adam optimizer with a learning rate of $10^{-3}$ that we drop to $10^{-4}$ halfway through training, if not stated otherwise. Once we have approximated the approximate Koopman operator, its spectral decomposition yields the Koopman eigenvalues and eigenvectors.

Turning to time-integration (prediction), we consider two distinct approaches, which we denote \KDLoo (Observable-Only evolution)  and \KDLso (State-Observable alternation), or \KDLAoo and KDLA$_{\textrm{so}}$ if we are finding $\Kapp$ with automatic differentiation. Note that in all cases, the first $n$ of the observables will always be chosen to be the state variables.
In \KDLoo, time evolution is entirely performed in the observable space. This approach maintains the linearity of the true Koopman operator framework. 
In \KDLso, each time step is performed in the space of observables, but at the end of the step, mapping back to the state is performed. Then at the beginning of the following step, the new state is mapped to the space of observables. Because of the application of the nonlinear functions $\boldsymbol{\chi}(\mathbf{x})$ and $\boldsymbol{\tilde{\chi}}(\mathbf{\psi})$, this approach is no longer linear.  Both algorithms for time evolution are presented in Algorithms 1 and 2, respectively. 
As we describe in Section III, the algorithm implemented by \citet{Qianxiao} was \KDLso, although the algorithm they described was \KDLoo. Results reported in this work correspond to \KDLAoo unless stated otherwise.

\begin{table}
% \caption{Time evolution algorithms}
\begin{minipage}{\textwidth} % Start a minipage to keep the algorithms together
% First Algorithm
\begin{algorithm}[H]
    \caption{: Time evolution in \KDLoo}\label{alg:second}
    \begin{algorithmic}
        \State $\boldsymbol{\psi} (t)=\boldsymbol{\chi}( \mathbf{x} (t)) $  \Comment{Given an IC in the state space, map to observable space}
        \State{$i=0,1 \ldots N_t$} \Comment{$N_t$ is the number of time steps}
        \For {$t_i$=$i \delta t + t_0$} 
        \State$\boldsymbol{\Psi} (t_i +\delta t) = \Kapp \boldsymbol{\Psi} (t_i)$  \Comment{Time evolution in the observable space }
        \EndFor
        \State $\tilde{\boldsymbol{\psi} } \gets [\boldsymbol{\Psi}(t_1),\boldsymbol{\Psi}(t_2) \cdots]$  \Comment{Matrix of predictions of observables }

        \State $\tilde{\mathbf{x}} (t) = \tilde{\boldsymbol{\chi}} (\tilde{\mathbf{\boldsymbol{\psi}} }(t))$  \Comment{Map back to full space }    
    \end{algorithmic}
\end{algorithm}

% Second Algorithm
\begin{algorithm}[H]
    \caption{: Time evolution in \KDLso} \label{table_time}
    \begin{algorithmic}
        \State{$i=0,1 \ldots N_t$} \Comment{$N_t$ is the number of time steps}
        \For {$t_i$=$i \delta t + t_0$} 
        \State $\boldsymbol{\Psi} (t_i)=\boldsymbol{\chi}( \mathbf{x} (t_i)) $  \Comment{Given an IC in the state space, map to observable space}
        \State$\boldsymbol{\Psi} (t_i +\delta t) = \Kapp \boldsymbol{\psi} (t)$  \Comment{Time evolution in observable space }
        \State $\mathbf{x} (t_i +\delta t) = \tilde{\boldsymbol{\chi}} (\boldsymbol{\psi} (t_i +\delta t) )$  \Comment{Map back to state space }
        \EndFor
        \State $\tilde{\mathbf{x} } \gets [\mathbf{x}(t_1),\mathbf{x}(t_2) \cdots]$  \Comment{Matrix of predictions }
    \end{algorithmic}
\end{algorithm}
\end{minipage} 
\end{table}

\begin{table}
\caption{ Architecture and parameters utilized in the studies of this paper for \KDLAoo.}
\centering
\setlength{\tabcolsep}{10pt} 
\begin{tabular}{ll*{4}{c}r}
\hline
Case & Shape & Activation & Dictionary elements \\
\hline
Duffing oscillator & 2/100/100/100/102 & ELU/ELU/ELU/ELU & 102 \\
Rossler attractor & 3/100/100/100/103 & ELU/ELU/ELU/ELU & 103\\
Flow past cylinder & 3/100/100/100/103 & ELU/ELU/ELU/ELU & 103\\
% Kuramoto model& 50/100/100/100/150 & ELU/ELU/ELU/ELU& 150\\
Burgers equation & 64/100/100/100/164 & ELU/ELU/ELU/ELU& 164\\
KSE  travelling wave & 64/100/100/100/164 & ELU/ELU/ELU/ELU& 164\\
KSE beating travelling wave & 64/100/100/100/114 & ELU/ELU/ELU/ELU& 114\\
KSE with chaotic dynamics & 64/250/250/250/214 & ELU/ELU/ELU/ELU& 214\\
Stuart-Landau equation & 1/50/50/50/26 & ELU/ELU/ELU/lin& 26\\
\label{Table_NN}
\end{tabular}
\end{table}

\section{Results}  
This section presents our \KDLAoo approach to approximate the Koopman operator on various dynamical systems of increasing complexity, including the low-dimensional state unforced Duffing oscillator and  
complex higher dimensional systems (i.e. discretized partial differential equations such as the Burgers equation and the Kuramoto-Sivashinsky equation). We also include results from \ALrevise{the state-observable alternation approach and the} state-space approach (NODE \cite{chen2019neural}). \ALrevise{Here we emphasize that the \KDLAoo approach is the only linear time evolution approach. As such, the nonlinear methods, state-observable alternation and the neural ODEs, provide a baseline for the best we might ever expect the \KDLAoo approach to perform.}

\subsection{Unforced Duffing oscillator \label{duffing_section}}

We apply our framework to the unforced Duffing equation, which describes an oscillator with a cubic nonlinearity.
The nondimensional equations of motion  are expressed as
\begin{equation}
\begin{aligned}
\dot{x}_1 &= x_2 \\
\dot{x}_2 &= -\lambda x_2 - x_1(\beta + \alpha x_1^2)
\end{aligned}
\end{equation}
where we will consider $\lambda=0.5$, $\beta=-1$ and $\alpha=1$  \citep{eDMD,Qianxiao}.  The dynamics are described by two stable spirals at $x_1 =\pm 1$
with $x_2=0 $, and a saddle at $x_1=x_2=0$; 
\textcolor{black}{thus, every initial condition, except for those on the stable manifold of the saddle, is drawn to one of these spirals.}
Overall, the Duffing oscillator is a challenging example for EDMD, but it is also a useful test case to compare the performance of our method with respect to other techniques  \cite{Otto_koopman,Qianxiao,eDMD}.

\begin{figure}
\begin{center} 
\begin{tabular}{ccc}
\includegraphics[width=0.36\linewidth]{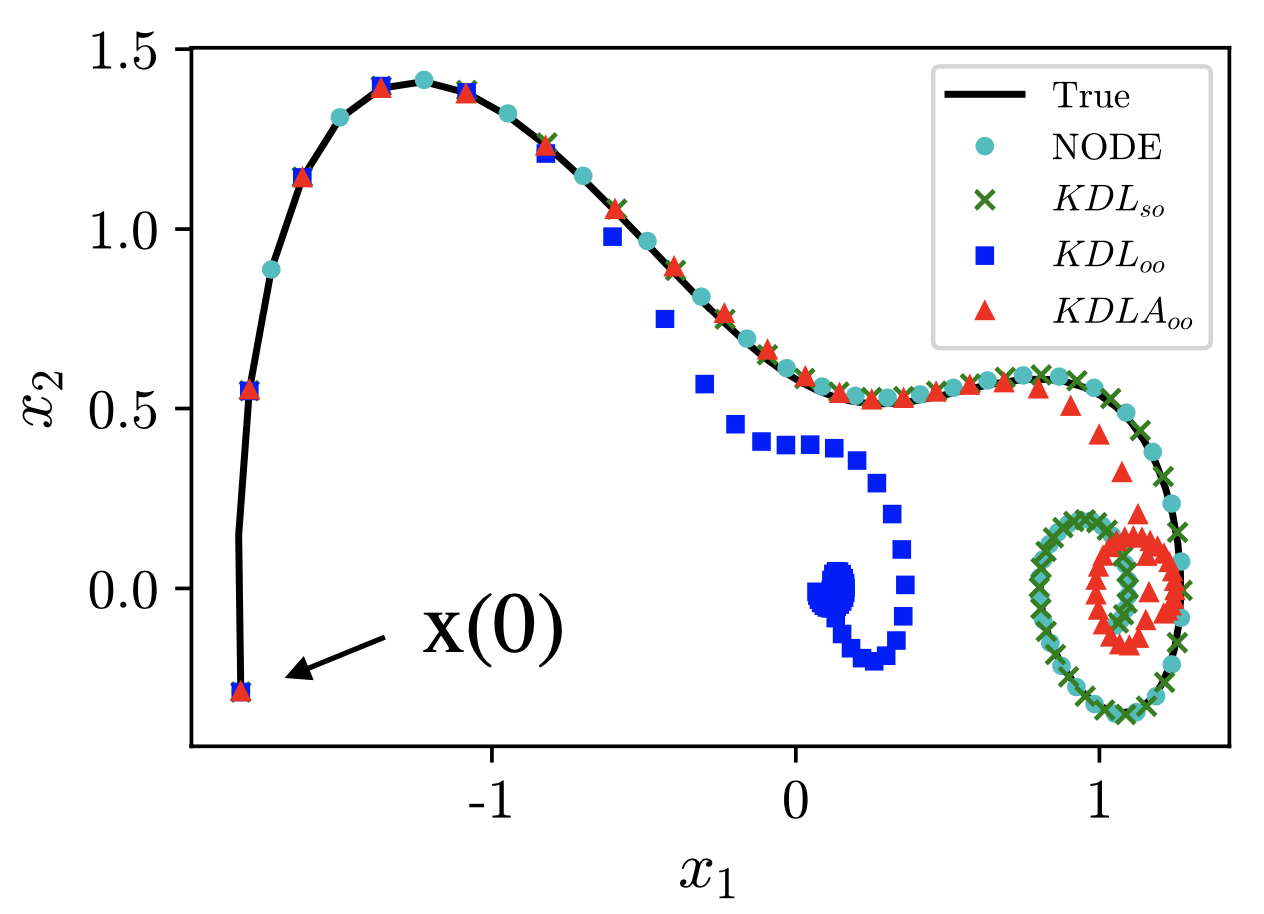}&
\includegraphics[width=0.29\linewidth]{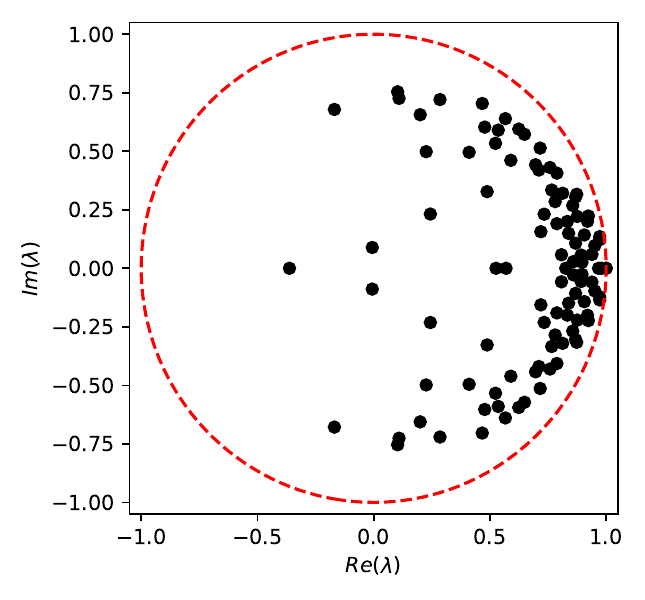}&
\includegraphics[width=0.35\linewidth]{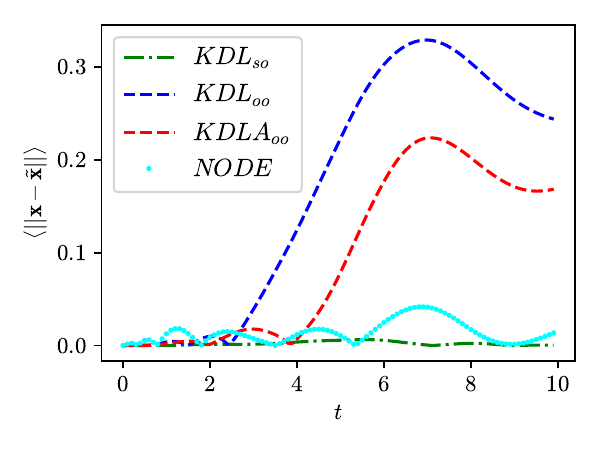}\\
(a) & (b) & (c) \\
\end{tabular}
\begin{tabular}{c}
\includegraphics[width=\linewidth]{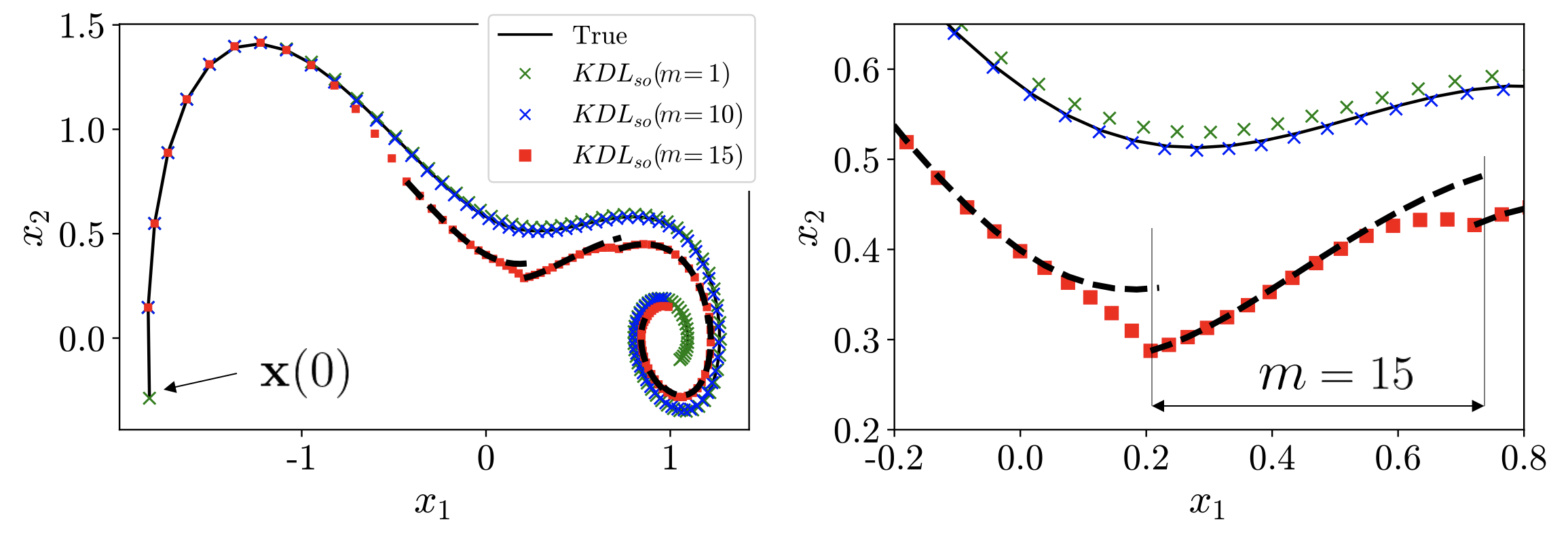}\\
(d)\\
\end{tabular}
\begin{tabular}{c}
\includegraphics[width=0.88\linewidth]{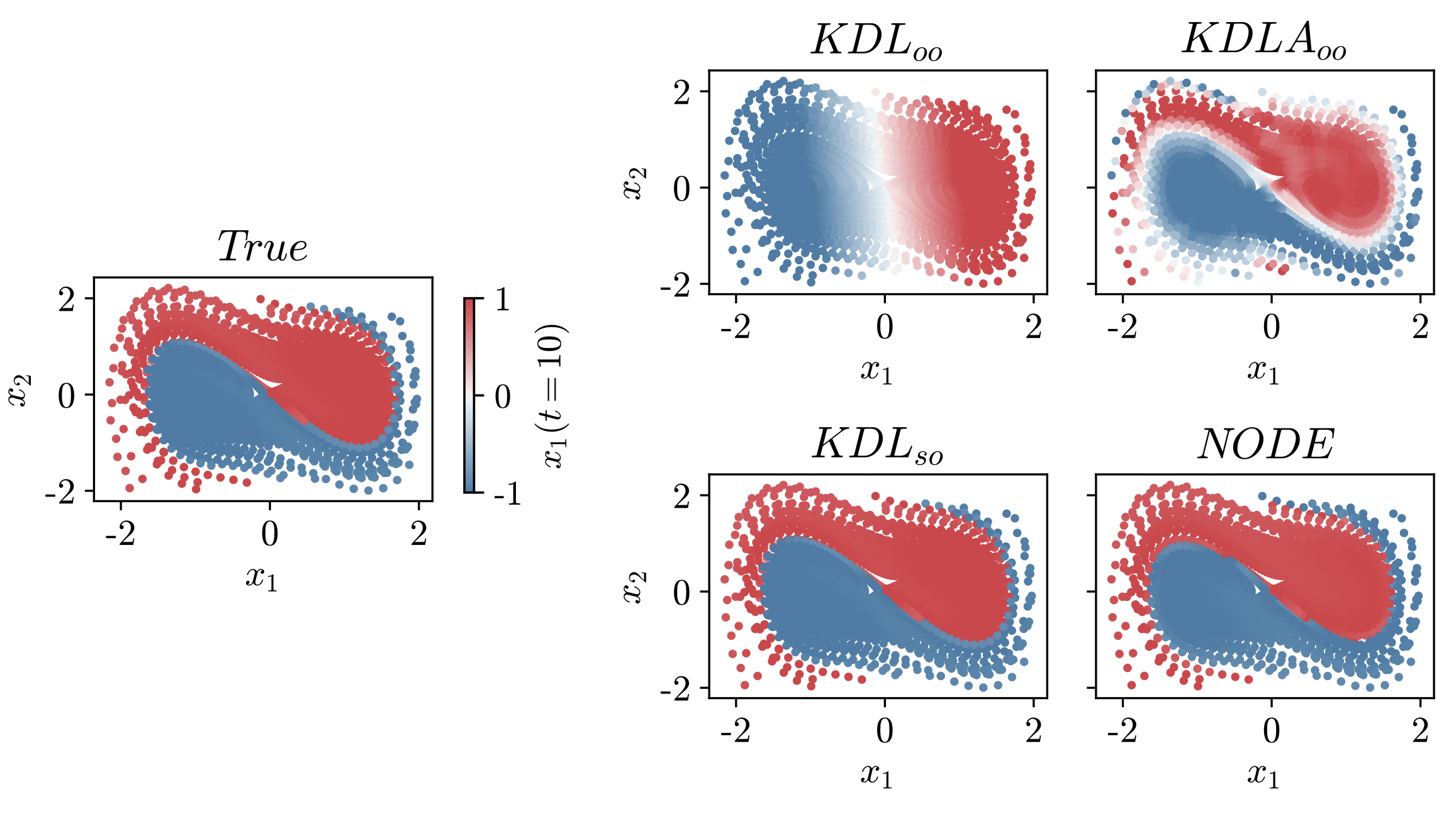}\\
(e)\\
\end{tabular}
\end{center} 
\caption{\label{duff}
\textcolor{black}{ Predictions for the Duffing oscillator with $\lambda=0.5$, $\beta=-1$ and $\alpha=1$.}
Panel (a) shows the trajectories reconstructed from 
\KDLso, \KDLoo, \KDLAoo, and the true data. 
Panel (b) shows the eigenvalues of the Koopman operator for \KDLAoo. Panel (c) shows the short-time error  computed by taking the
 mean squared difference of 1000 different initial conditions between the exact trajectory and the different models as a function of time.
\textcolor{black}{Panel (d) displays the temporal evolution of the same initial condition for different variants  \KDLso as a function of the number of steps ($m$) that the system has remained within the observable space. To provide a more detailed view, a magnified version of this panel is presented on the right. The dashed lines correspond to the new exact trajectories after reintroducing the state variables through the dictionary.}
Panel (e) shows the long-time predictions in a $x_1$-$x_2$ plane, in which each initial condition is coloured by its final value of $x_1$ at $t=10$. 
 } 
\end{figure}

The dataset consists of   100 random initial conditions of the state  $\textbf{x}$ where $x_1,x_2 \in [-2, 2]$ that forms a large enough 
\MDGrevise{region in state space}  to capture the underlying dynamics of the two basins of attraction. 
We store data at intervals of $\delta t =0.1$, and 
each initial condition is forecast to $100$ time steps.
We create  100 dictionary elements in addition to the  state  $\textbf{x}$; thus, the vector of observables $ \mathbf{\Psi }$ is 102-dimensional. 

For the \KDLso and \KDLoo  frameworks, \textcolor{black}{we follow the hyperparameters proposed by \citet{Qianxiao}; thus,}
the architecture of the neural network consists of three hidden layers with 100 nodes per layer using $\tanh$ as the activation function, and the output layer forms 22 trainable dictionary elements, \textcolor{black}{along with a constant element and the state variables} (i.e., a total number of 25 dictionary elements).
The training consists of  3000 epochs with a batch size of 5000, 
a learning rate of $10^{-4}$, and a value of $0.1$ for the Tikhonov regularizer.
%We remark that the choices for the network architecture and hyperparameters follow closely those described by \citet{Qianxiao}.
\textcolor{black}{As noted above, for \KDLAoo, we use the hyperparameters given in Table \ref{Table_NN}.}
\textcolor{black}{The appendix A  presents a comparison of different architectures for \KDLoo, and the rationale behind selecting the architecture proposed by \citet{Qianxiao}, as opposed to the architecture employed  for \KDLAoo. }
For the state space approach, the training data is the same as described above \ALrevise{(i.e. snapshots of the two-dimensional state $[x_1,x_2]^T$)} and the architecture of the neural network used for the neural ODE  consists of two hidden layers with 200 nodes per layer with sigmoid activation functions except for the last layer, which is linear. We keep this architecture for all the NODE and \KDLoo presented in this work, unless stated otherwise.

Figure \ref{duff}a depicts the state space trajectories in the $x_1$-$x_2$ plane for the same initial condition using four distinct methods: \KDLso, \KDLoo, \KDLAoo, and NODE. Our approach, \KDLAoo, outperforms the predictions generated by \KDLoo, as it is capable of predicting the true dynamics for longer times. Notably, the trajectories produced by \KDLso and NODE exhibit close agreement with the ground truth, underscoring their accuracy. 
Figure \ref{duff}b shows the eigenvalue spectrum of the Koopman operator for \KDLAoo. 
To illustrate the performance of all these models, we present short-time tracking results in figure \ref{duff}c. It displays  the normalized ensemble-averaged tracking error (i.e., $\langle||\textbf{x} - \tilde{\textbf{x}} ||\rangle $) up to $t=10$
calculated in the test data set (e.g., 1000 initial conditions not seen during training).
At short times the error is small for all approaches, but \KDLAoo  exhibits significantly increased predictive capabilities compared to \KDLoo. 
For \KDLoo and \KDLAoo, after $t>4$, there is a rise in the error until $t=7$, after which  \KDLAoo dips below \KDLoo, suggesting that our approach lands closer to the fixed points. We remark that \KDLso and NODE lead to small errors.  
This short-time tracking also agrees with the method presented by \citet{Otto_koopman} and standard EDMD (see their figure 2a).

As we pointed out in Section \ref{koopman_approx}, the  \KDLso framework corresponds to iterating between the state and observable spaces at each time interval.
Now, our focus is to underscore the vital role played by this iterative process  in enhancing the predictive capabilities.
We will examine \KDLso  by introducing a variant defined by  the number of time intervals $m$ that the system spends in 
the observable space before returning to the state space. In the limit of $m=1$, \KDLso corresponds to the work presented by \citet{Qianxiao}, while   $m\rightarrow\infty$ corresponds to a `pure' Koopman approach, where all the evolution occurs solely in the space of observables. We 
Figure \ref{duff}d presents the temporal evolution on the $x_1$-$x_2$ plane, starting from an initial condition converging towards the spiral with $x_1=1$ as a function of $m$. When $m$ is small, the iterative process between the state and observable spaces yields accurate predictions as the mapping between spaces effectively corrects the predictions to match the true dynamics.  However, as $m$ grows larger, the iterative process struggles to produce accurate predictions over longer time intervals. This limitation arises due to the inability of the approximated dictionary and Koopman operator to adequately capture the system's dynamics -- essentially, small errors accumulate. 
To further emphasize this point, consider the case with $m=15$ presented in the right panel of figure \ref{duff}d. After $15$ time intervals within the observable space, the Koopman predictions start to deviate from the real solution.
 But, when the state is taken out and reintroduced to the observable space, the temporal evolution aligns with the exact solutions of the  new initial conditions (the exact solution of these new initial conditions is depicted by the black dashed lines in figure \ref{duff}d). This  highlights that predictions are enhanced when $m$ is smaller. \textcolor{black}{The success of the back-and-forth approach is very promising, although not fully understood. Further analysis is ongoing, but we can confirm that the alternation between state- and observable-spaces introduces nonlinearity, disrupting the advantages of a linear system.}

Next, we investigate the long-time predictive capabilities of the different methods. Figure \ref{duff}e shows a phase diagram with 1000 random initial conditions coloured by their final value of $x_1 (t=10$).
For any initial condition $\textbf{x}(0)$, the value of $x_1$ will ultimately settle into one of two possible states: either 0 or $\pm 1$ (as can be seen in the phase diagram of the true dynamics).
When examining long-term predictive capabilities for `pure` Koopman approach, \KDLAoo stands out with respect to \KDLoo for its ability to better capture effectively the system's long-time dynamics. \KDLoo  fails to predict the basins of attraction, aligning with our previous observation of the Koopman operator's poor approximation in this approach. \ALrevise{Thus, our approach to the optimization procedure results in observables and an approximation of the Koopman operator that substantially improves the long-time predictive capabilities in comparison to the iterative procedure used in \KDLoo . When the state-observable alternation approach is taken}
 for \KDLso with $m=1$, 
the long-term tracking is comparable to the true trajectories \textcolor{black}{(similar findings with EDMD were reported by \citep{Junker})}.
Finally, when comparing the state space approach (NODE) with the function space approaches, we observe better predictive capabilities for the state space approach for short and long-time tracking relative to \KDLoo and \KDLAoo, and comparable performance relative to \KDLso.

From here, our emphasis will be on assessing the performance of various methods within a 'pure' Koopman approach. We will refrain from showcasing additional predictions of \KDLso, as it no longer reflects a linear evolution, thereby diminishing its alignment with the true Koopman operator.

\subsection{ \textcolor{black}{R\"{o}ssler} attractor }

Next, we try our methodology in the \textcolor{black}{R\"{o}ssler} equation, which exhibits chaotic dynamics in a three-dimensional system of ordinary differential equations. This equation is  
\begin{equation}
\begin{aligned}
\dot{x}_1 &= -x_2 - x_3 \\
\dot{x}_2 &= x_1 + a x_2 \\
\dot{x}_3 &= b + x_3 (x_2 - c)
\end{aligned}
\end{equation}
where $a$, $b$, and $c$ are constants.
We chose values $a=b=0.1$ and $c=9$, where the dynamics form a  `sparse chaotic attractor'. 
The dataset consists of one long trajectory up to $t=1000$, with $\delta t=0.1$. We remove the short transient dynamics, so the data live   completely on the attractor.
We create  100 dictionary elements in addition to the state  $\textbf{x}$; thus, the set of observables $ \mathbf{\Psi }$ is 103.

\begin{figure}
\begin{center} 
\begin{tabular}{ccc}
\includegraphics[width=0.33\linewidth]{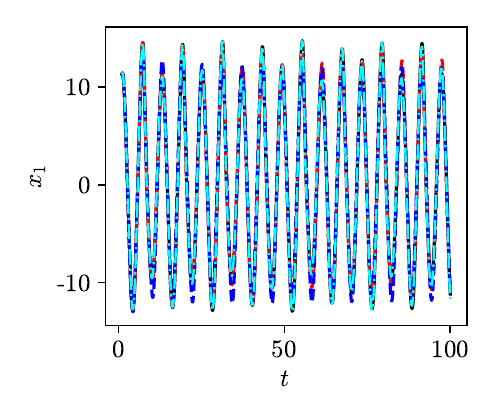}&
\includegraphics[width=0.33\linewidth]{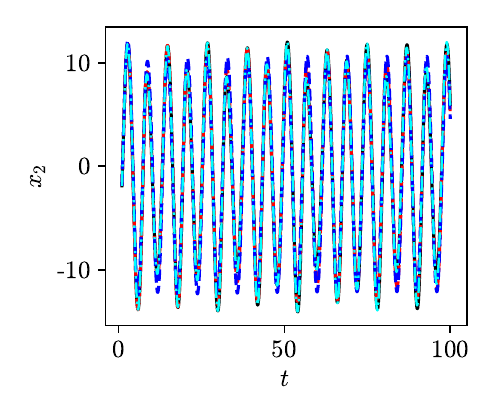}&
\includegraphics[width=0.33\linewidth]{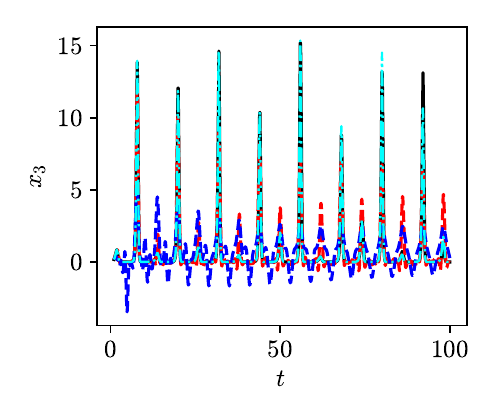}\\
  &  (a) & \\
\includegraphics[width=0.34\linewidth]{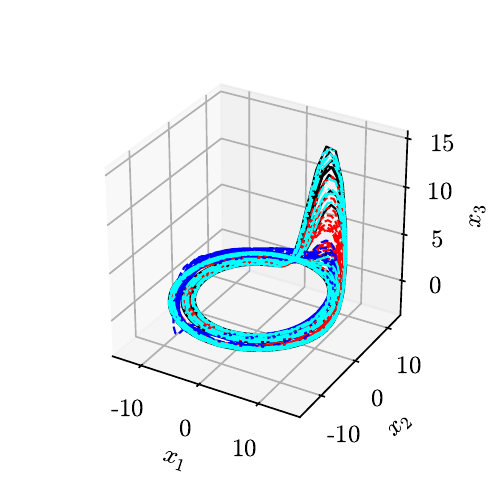}&
\includegraphics[width=0.32\linewidth]{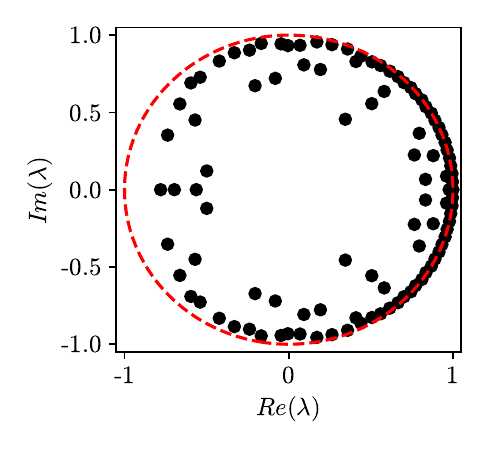}&
\includegraphics[width=0.32\linewidth]{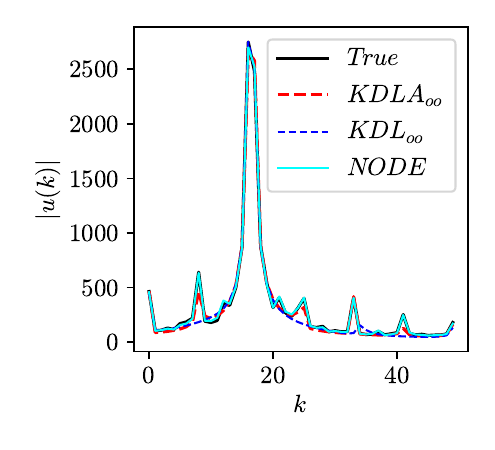}\\
 (b) &  (c) & (d)\\
\end{tabular}
\end{center} 
\caption{\label{Rossler}
Predictions for the chaotic \textcolor{black}{R\"{o}ssler} attractor with $a=0.1$, $b=0.1$, and $c=9$. Panel (a) shows the temporal evolution of the three components of $\textbf{x}$ for\KDLoo, \KDLAoo and neural ODE. Panels (b-d) show a 3D representation of the \textcolor{black}{R\"{o}ssler} structure,  the eigenvalues of the Koopman operator for the \KDLAoo, and the power spectrum of the model's predictions and true data, respectively. 
} 
\end{figure}

Figure \ref{Rossler}a shows the temporal prediction of one initial condition that starts on the attractor (but is not seen by the model during training). We observe good agreement with respect to the true data in relation to $x_1$ and $x_2$, but for $x_3$, \KDLAoo  is underpredicting the amplitude of the spikes at long times. Figure \ref{Rossler}b shows a three-dimensional representation of the trajectory for the true and data-driven models.  Figure \ref{case3_lush}c shows the {eigenvalues} from the Koopman spectrum of the \KDLAoo; most are on the unit circle. Finally, figure \ref{Rossler}d shows the power spectra (that are calculated from the discrete Fourier transform DFT) for the true data and the model \ALrevise{prediction. Here we note that this power spectrum was computed only over the time range covered by the model trajectory in Fig.\ \ref{case3_lush}b. The subsequent power spectrums will also be computed for the example trajectories shown.}  It shows that the structure of the data is dominated by a single length scale with wavenumber  $k$ around 15 to 17. Our method's predictions are capable of capturing the structure of the data. 
For this case \KDLAoo predicts negative values for $x_3$, and in general underestimates the peaks in $x_3$. However, 
 \KDLAoo outperforms \KDLoo with a better prediction of the values of $x_3$ .
Finally, for this case, NODE outperforms the Koopman's predictions capturing perfectly the dynamics of the attractor, as it can be observed in figure  \ref{Rossler}a,d.

\textcolor{black}{ Given the inherently chaotic nature of the system, we expect at best a reasonable short-term trajectory tracking before the underlying dynamics deviate significantly from the original solution. Figure 3a  illustrates this phenomenon, where the NODE model adeptly captures the system dynamics within the displayed time span. However, the performance of  \KDLAoo becomes noticeably less robust, diverging notably after $t>25$, and finally,  \KDLoo exhibits a swift divergence from the true solution, indicating a substantial inadequacy in approximating the Koopman operator. While NODE demonstrates commendable short-term accuracy, the divergence in  \KDLAoo and \KDLoo  serves as a compelling indicator of the challenges associated with accurately representing the intricate dynamics of chaotic systems. }

\textcolor{black}{The decrease in the energy  is due to the eigenvalues lying inside the unit circle. Several approaches have been proposed   to address this issue by explicitly constraining the eigenvalues to unity (such as, adding a  soft constraint for orthogonality in the loss function or the implementation of the Procrustes problem \citep{colbrook_ayton_2023,procrustes}). Incorporating this framework to the systems of this study is the subject of future work.}

\subsection{ \textcolor{black}{Three-dimensional model for flow past a cylinder  }}

\begin{figure}
\begin{center} 
\begin{tabular}{cc}
\includegraphics[width=0.4\linewidth]{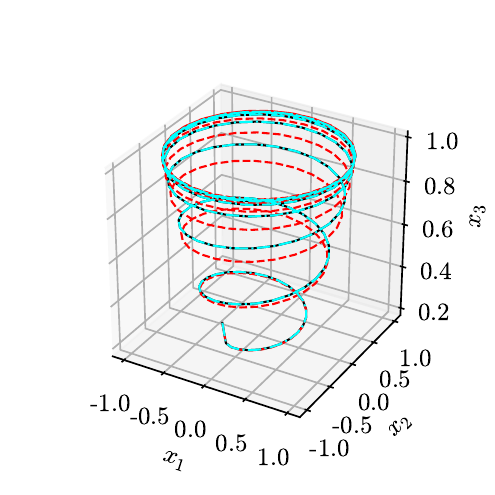}&
\includegraphics[width=0.4\linewidth]{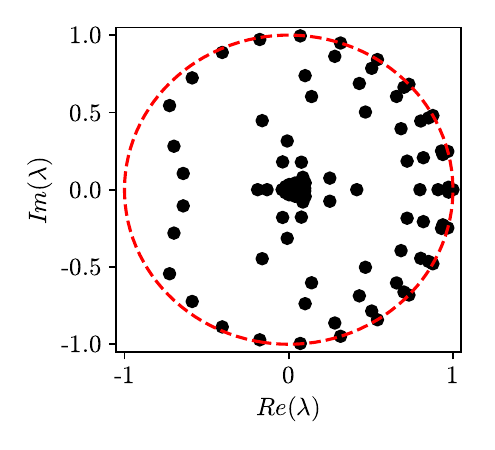}\\
(a) & (b) \\
\includegraphics[width=0.4\linewidth]{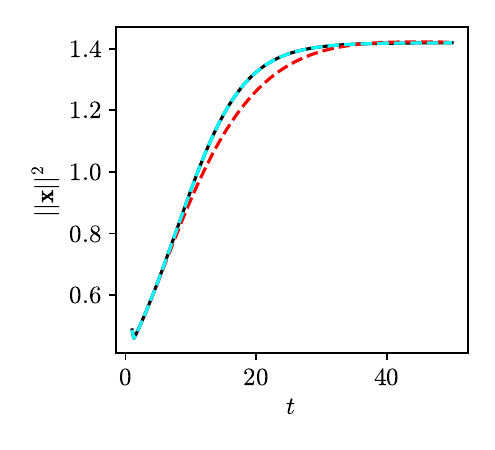}&
\includegraphics[width=0.4\linewidth]{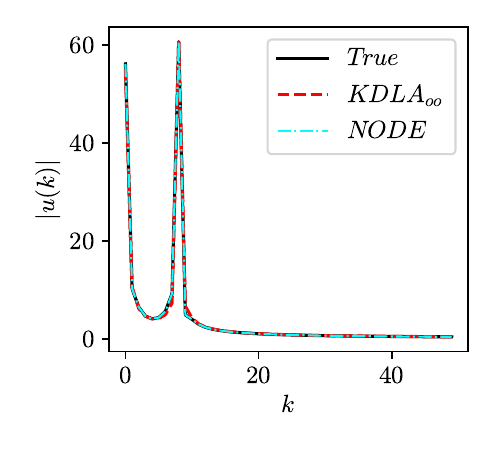}\\
(c) & (d) \\
\end{tabular}
\end{center} 
\caption{\label{case3_lush}
Predictions for \textcolor{black}{a three-dimensional model for} flow past a cylinder. Panel (a) shows a 3D representation of the trajectory for the true data, \KDLAoo and NODE for an initial condition that starts off the attractor, and eventually is attracted  onto the limit circle.
Panels (b-d) show the eigenvalues of the Koopman operator for the \KDLAoo, the energy of the system and the power spectrum for the model's predictions and true data, respectively. } 
\end{figure}

We now apply our framework to study the flow dynamics past a cylinder  by considering a three-dimensional system of ordinary differential equations. In the fluid dynamics community, this case has attracted great attention to illustrate modal decomposition  techniques \citep{Loiseau,dmd_book,lando,Lusch,noack_2003}.
\citet{noack_2003} showed that the \textcolor{black}{transient dynamics of a cylinder wake } with $Re=DU/\nu=100$ (e.g., where $D$ is the diameter of the cylinder, $U$ is the free-stream velocity and $\mu$ is the kinematic viscosity)
can be represented by  a 
three-dimensional system:
\begin{equation}
\begin{aligned}
\dot{x}_1 &= \mu x_1 - \omega x_2 + A x_1 x_3  \\
\dot{x}_2 &= \omega x_1 + \mu x_2 + A x_2 x_3 \\
\dot{x}_3 &= -\lambda (x_3 - x_1^2 - x_2^2)
\end{aligned}
\end{equation}
where $\mu=1/10$, $\omega=1$, $A=-1/10$ and $\lambda=10$. This model is characterized by an unstable fixed point at the origin corresponding to 
\MDGrevise{steady flow}, and a stable limit cycle\MDGrevise{, capturing oscillatory vortex-shedding,} that is asymptotically and globally stable.

The dataset consists of  100 initial conditions   $\textbf{x}$ where $x_1 \in [-1.1, 1.1]$,
$x_2 \in [-1.1, 1.1]$ and
$x_3 \in [0, 2.42]$.
\ALrevise{The selected boundaries ensure that we include initial conditions off of the slow manifold, which is characterized by the paraboloid $x_3=x_1^2 +x_2^2$.}
We store data at intervals of $\delta t =0.25$ (to be able to capture the rapid transit dynamics, $\delta t$ should be reduced further), and 
each initial condition is forecast on time up to $t=50$ time units.
We create  100 dictionary elements in addition to the state  $\textbf{x}$; thus, the set of observables $ \mathbf{\Psi }$ is 103. 

Figure \ref{case3_lush}a shows the temporal prediction of one initial condition that starts away from the attractor, and the system evolves towards the low-dimensional paraboloid manifold.
Figure \ref{case3_lush}a shows that \KDLAoo is capable of predicting the transient dynamics that are rapidly attracted toward the limit cycle via the paraboloid. 
Figure \ref{case3_lush}b shows the eigenvalues from the Koopman spectrum of the \KDLAoo. We observe that some eigenvalues are on the unit circle which  indicates that the dynamics will take a long time to decay.  Finally, figure \ref{case3_lush}c and d show the energy and the power spectra for the true data and the models, respectively. For this case, both the state space and function space approaches, \KDLAoo,  lead to good predictions of the dynamics.

\subsection{Viscous Burgers equation}

To test our method on data generated by a partial differential equation  we consider
 the viscous Burgers equation that describes the motion of fluid in one dimension (e.g., it can be seen as a simplified version of the Navier-Stokes equation).  
 The  Burgers equation is expressed as
\begin{equation}
\centering
u_t = \nu u_{xx}- u u_x
\label{burger_eq}
\end{equation}
where $u(x,t)$ is the velocity in $x \in [-1,1 ]$ and time $t > 0$, and $\nu$ is the kinematic viscosity ($\nu=0.01$). The subscripts $x$ and $t$ stand for partial derivatives with respect to space and time. The first term of the right-hand side represents the diffusive transport due to viscosity, while the second term stands for the convective transport of fluid.
The selection of $\nu=0.01$ is to reduce the effect of the linear contribution $\nu u_{xx}$ with respect to the nonlinear component $- u u_x$  

To generate the training data, we  solve the previous  equation  using the spin operator in the Chebfun package \citep{Chebfun}, we set the initial conditions as
\begin{equation}
\centering
u(x,0) =3 A_1 \textrm{sech}^2 (3 \sin (\pi(x-2s_1)+5 A_2 \textrm{sech}^2(3 \sin(\pi(x-2s_2))
\end{equation}
where $A_{1,2}$ and $s_{1,2}$ stand for constants randomly distributed in the interval $[0,1]$. 

The training data consists of 20 \textcolor{black}{random initial conditions} integrated up to $t=20$,  
with a discrete-time spacing $\delta t=0.1$.  We discretize the PDE on a grid, $\Delta x$, with 64 points, making the state space 64-dimensional.
We create  100 dictionary elements  in addition to the state  $\textbf{x}$; thus, the set of observables $ \mathbf{\Psi }$ is 164. 

\begin{figure}
\begin{center} 
\begin{tabular}{cc}
\includegraphics[width=0.55\linewidth]{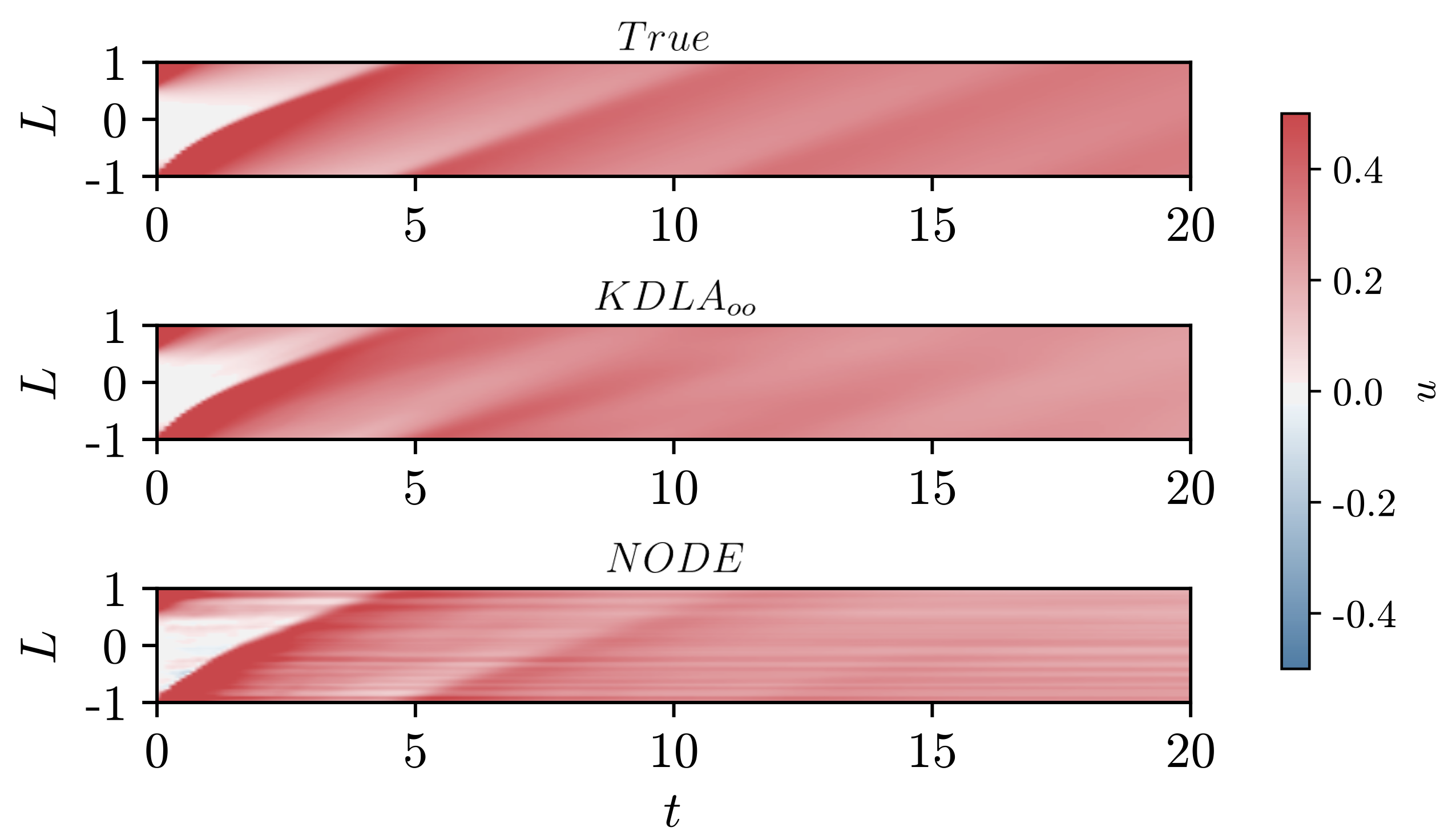}&
\includegraphics[width=0.45\linewidth]{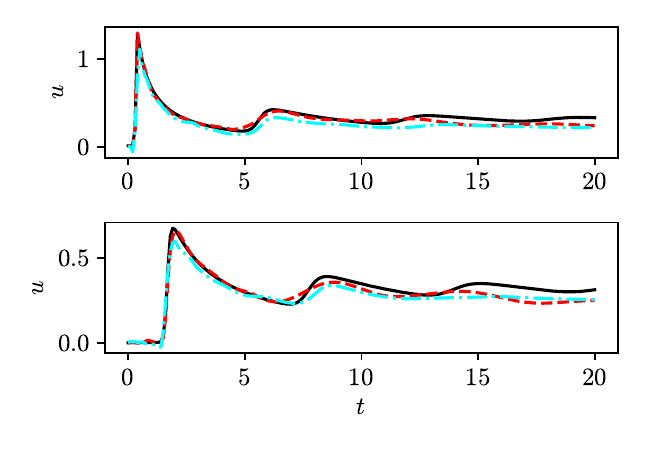}\\
(a) & (b) \\
\includegraphics[width=0.42\linewidth]{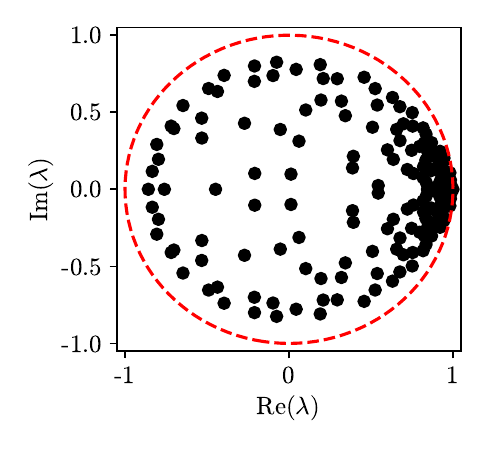}&
\includegraphics[width=0.42\linewidth]{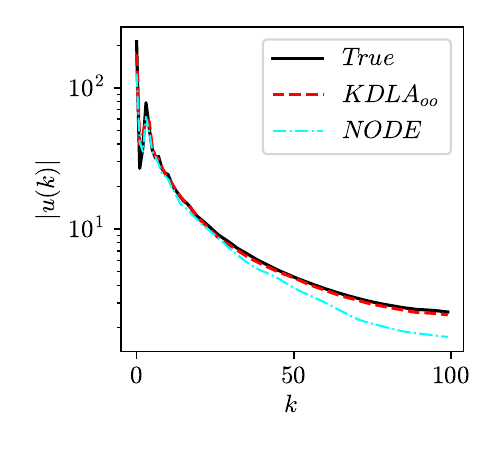}
\\  (c) &(d) \\
\end{tabular}
\end{center} 
\caption{\label{burger_fig}
Predictions for the viscous Burger's equation. Panel (a) shows a snapshot of the true  (top panel) and predicted dynamics for the \KDLAoo and NODE (middle and bottom panels, respectively). Panel (b) compares the value of $u(x,t)$ from the model's predictions and the true state when  \textcolor{black}{$u=(x=0,t)$, $u=(x=0.5,t)$,} corresponding to the top and bottom panels, respectively.  Panels (c)-(d) represent the eigenvalues of the Koopman operator for \KDLAoo, and  the power spectrum of the model's predictions and true data, respectively.
} 
\end{figure}

The top panel of figure \ref{burger_fig}a shows the true dynamics of a single trajectory, in which strong nonlinearities are seen at early times of the dynamics, and they decay as time evolves.  The middle and bottom panels of figure \ref{burger_fig}a represent the predictions  from the \KDLAoo and NODE approaches, respectively, both models  show a good agreement with respect to the true dynamics. \ref{burger_fig}b shows a comparison of the values of $u(x=0,t)$ and $u(x=0.5,t)$ between the true and predicted models. Figure \ref{burger_fig}c shows the eigenvalues of the { Koopman operator}. Figure \ref{burger_fig}d shows the power spectrum and demonstrates that the \KDLAoo can predict well the structure of the data.
\textcolor{black}{In the context of the neural ODE approach, \ALrevise{there is} a mismatch with the true data, primarily attributed to the amplification of high wavenumbers over time. This} \MDGrevise{behavior has been previously observed with neural ODE treatment of PDEs \cite{alec_chaos} and can be addressed by adding a linear damping term to the neural ODE right-hand side \citep{alec_Stabilized}. }

\subsection{Kuramoto-Sivashinsky Equation}

The one-dimensional Kuramoto-Sivashinsky equation (KSE) is a  dissipative partial differential equation that exhibits  spatiotemporal chaotic behavior:
\begin{equation}
u_t=-u_{xx}-u_{xxxx}-\frac{1}{2}uu_x~~, ~~~~~~~~~x  \in [-L/2, L/2]
\end{equation}
 on a periodic domain $u(x,t)=u(x+L,t)$, where $t$, and $x$ stand for the time and spatial coordinates, respectively. 
The nonlinear term (i.e., the third term of the RHS) drives the transfer of energy from low to high wavenumbers. \ALrevise{For the KSE, energy is injected into the system through the second derivative term and dissipated through hyperdiffusivity in the fourth derivative term.}
We \ALrevise{again} solve this \ALrevise{equation} using the spin operator in the Chebfun package \citep{Chebfun}.
The solver uses exponential time differencing with fourth-order stiff time-stepping \citep{Cox}.

We prepare our training data by computing a numerical solution on a grid of $64$, $\textbf{x} \in \mathbb{R}^{64}$.
We discard the initial transient dynamics, then we ensure that the set of training data has landed on the \ALrevise{long-time} attractor.
The state vector $\textbf{x}$ corresponds to the solution $u$ at the  grid points. 
The initial conditions considered in this study consist of finite Fourier series with distributed coefficients of equal variance \citep{Filip,lando}.

\subsubsection{Kuramoto-Sivashinsky  traveling wave  }

The Kuramoto-Sivashinsky equation with  $L=12$ leads to a traveling wave regime \citep{cvitanovic2005chaos}. 
The training data consists of one long trajectory integrated up to $t=10^4$, 
with a discrete-time spacing $\delta t=0.25$. 
We create  100 dictionary elements in addition to the state  $\textbf{x}$; thus, the set of observables $ \mathbf{\Psi }$ is 164. 

\begin{figure}
\begin{center} 
\begin{tabular}{cc}
\includegraphics[width=0.55\linewidth]{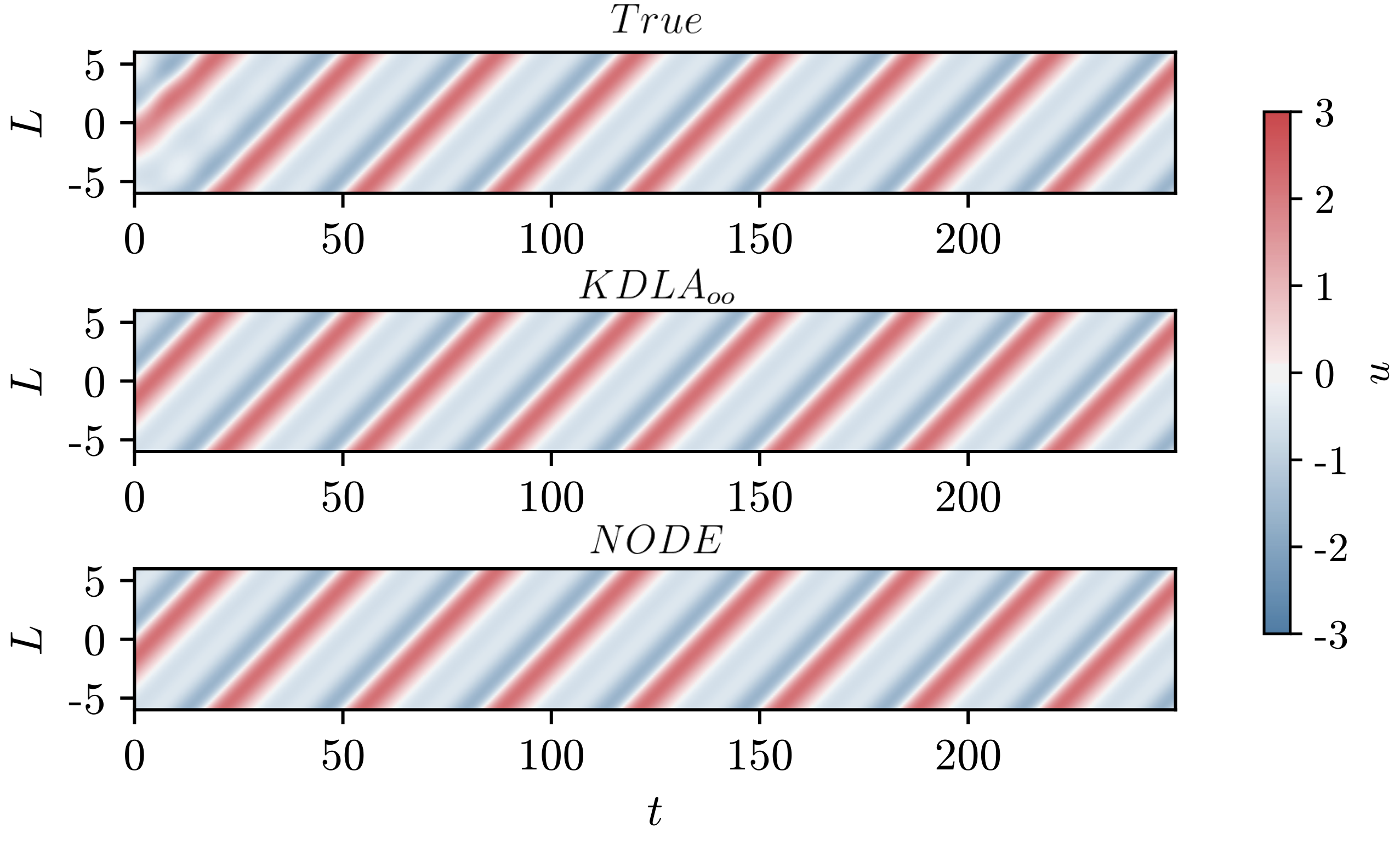}&
\includegraphics[width=0.45\linewidth]{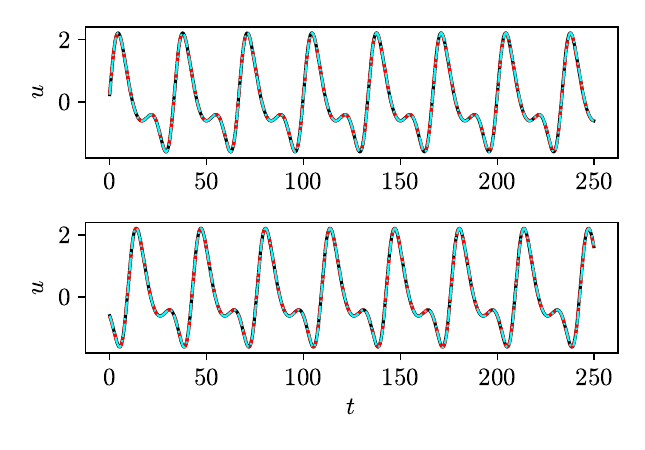}\\
(a) & (b)  \\
\end{tabular}
\begin{tabular}{ccc}
\includegraphics[width=0.33\linewidth]{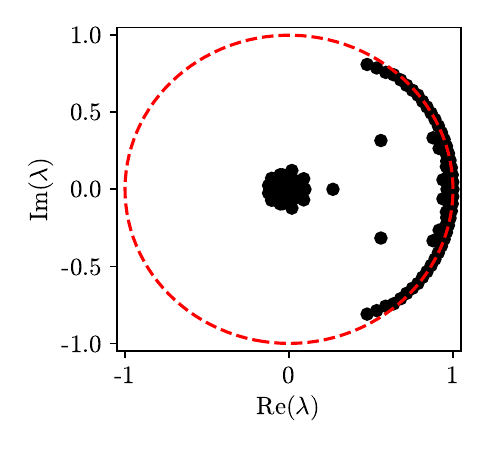}& 
\includegraphics[width=0.33\linewidth]{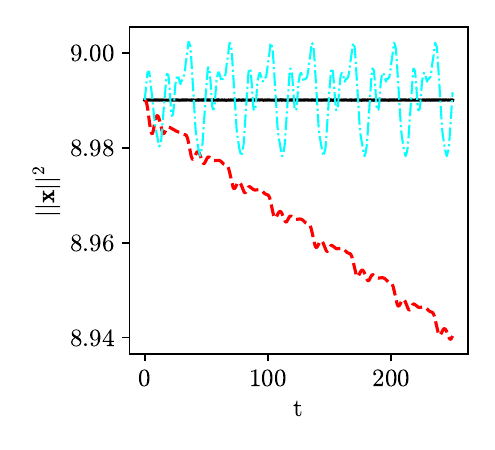}& 
\includegraphics[width=0.33\linewidth]{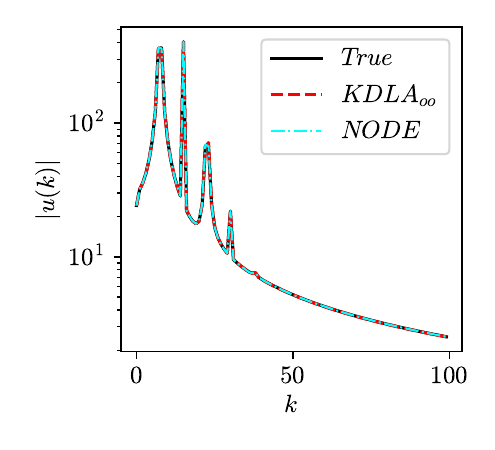}\\
  (c) & (d) & (e) \\
\end{tabular}
\end{center} 
\caption{\label{TW_KS}
Predictions for  the Kuramoto-Sivashinsky with traveling waves at $L=12$. 
Panel (a) shows a snapshot of the true  and predicted dynamics for the \KDLAoo and NODE, corresponding to top to bottom panels, respectively. Panel (b) compares the value of $u(x,t)$ from the model's prediction to the true state when  $u=u(0,t)$, $u=u(2.5,t)$, corresponding to the top and bottom panels, respectively.  Panels (c)-(e) represent the eigenvalues of the Koopman operator for the \KDLAoo, the system energy and the power spectrum of the model's predictions and true data, respectively.
} 
\end{figure}

The top panel of figure \ref{TW_KS}a shows the \ALrevise{true traveling wave dynamics.}
The middle and bottom panels show the trajectory that results from evolving the same initial condition from the top panel forward on time under our approximated Koopman operator via \KDLAoo, and using the NODE. 
Comparing the three panels, we see that predictions of both models result in excellent agreement, as can be observed in the good match of the values of the solution at different spatial locations over time displayed in figure \ref{TW_KS}b (e.g., $u(x=0,t)$ and $u(x=0.5,t)$).
Figure \ref{TW_KS}c and d show the eigenvalues of the Koopman operator, and the temporal evolution of the ``energy" $||\textbf{x}||^2$ for the models and ground truth. For the \KDLAoo, a gradual energy decay of $0.5\%$ over a span of $250$ time units is observed. \ALrevise{This gradual decrease in energy is due to a small deviation of the eigenvalues from the unit circle.}
Figure \ref{TW_KS}e shows the power spectra for the true data and the model predictions up to $t=250$,
where it shows that \KDLAoo and NODE  are capable of capturing very accurately the structure of the data. 
Therefore, figure \ref{TW_KS} shows that both the state space and function space approaches lead to excellent agreement with the ground truth for traveling wave dynamics.

\subsubsection{Kuramoto-Sivashinsky beating traveling waves  }

Our next example also comes from the KS equation with $L=29.30$
that results in beating traveling waves at long times. 
The traveling period and beating period are incommensurate, making the orbit quasiperiodic. Note the separation in timescales: the two periods differ by a factor of nearly 200. This quasiperiodic orbit lives on a two-dimensional submanifold (a 2-torus) of $\mathbb{R}^{64}$.

Our training set consists of a single trajectory evolved up to $t=2000$ in a grid of  $=64$ points and $\delta t=0.05$.   
We create  50 dictionary elements in addition to the state  $\textbf{x}$; thus, the set of observables $ \mathbf{\Psi }$ is 114. 
For this case, we also include results from the \KDLoo in which the time integration is entirely performed in the Koopman space. For the \KDLoo, we  select 3 hidden layers with 100 neurons to make fair comparisons.

\begin{figure}
\begin{center} 
\begin{tabular}{cc}
\includegraphics[width=0.55\linewidth]{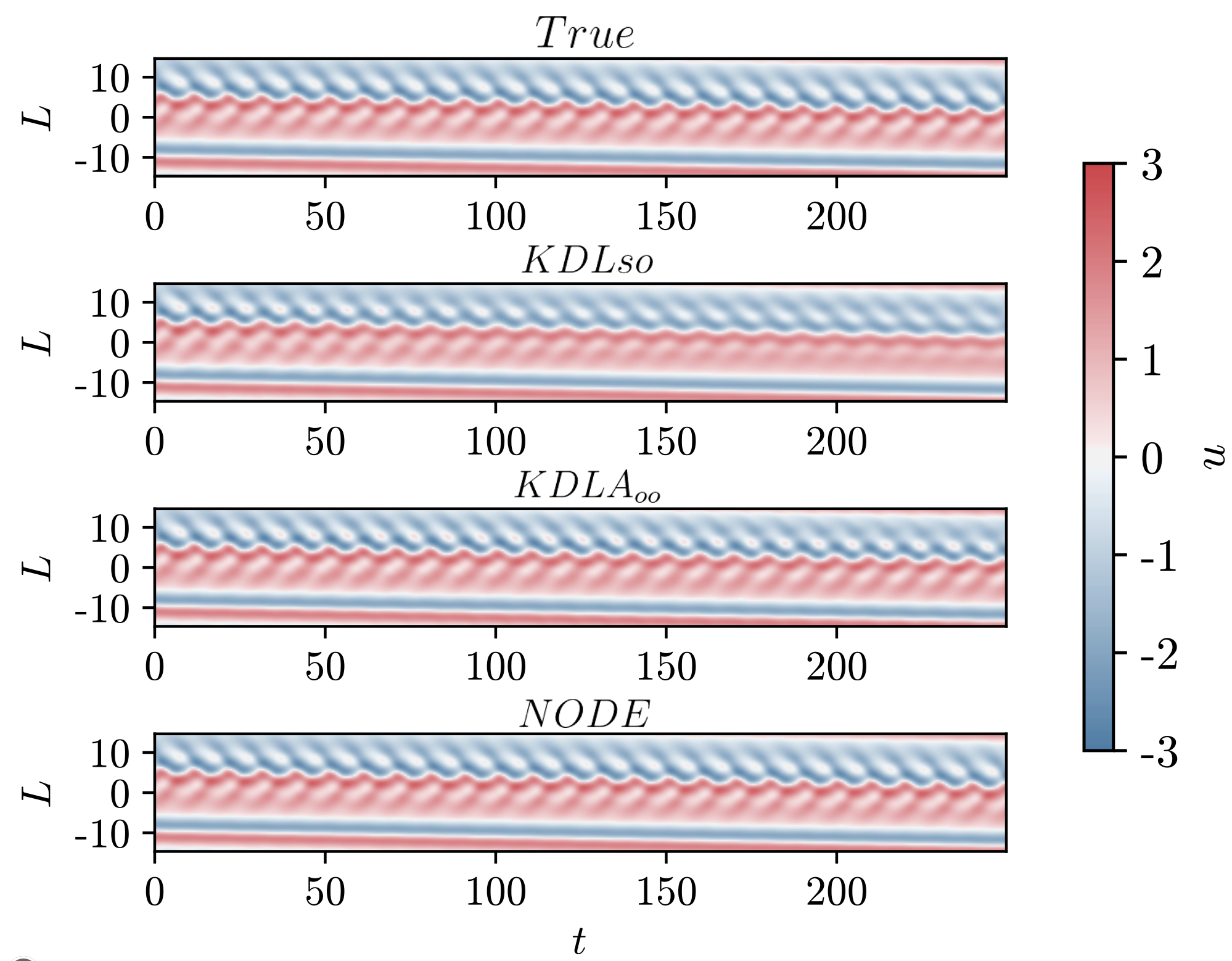}&
\includegraphics[width=0.45\linewidth]{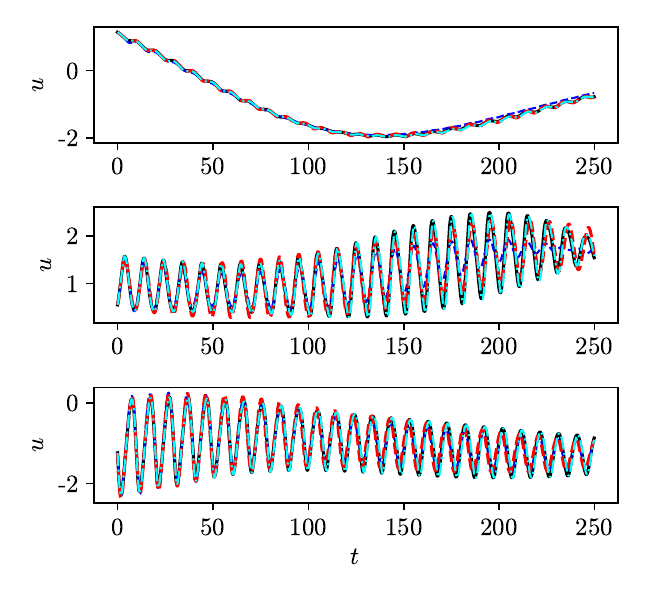}\\
(a) & (b)  \\
\end{tabular}
\begin{tabular}{ccc}
\includegraphics[width=0.33\linewidth]{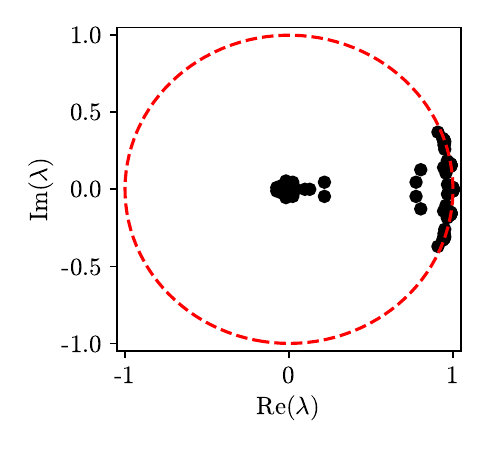} &
\includegraphics[width=0.33\linewidth]{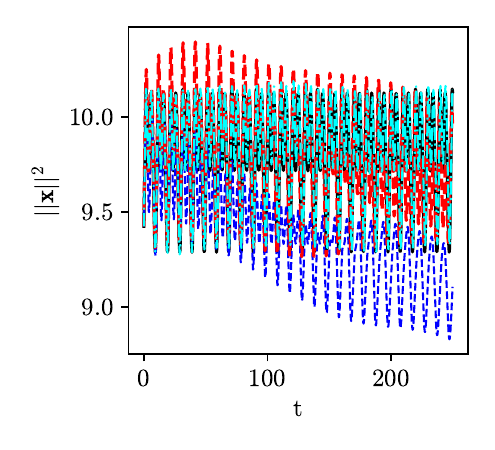} &
\includegraphics[width=0.33\linewidth]{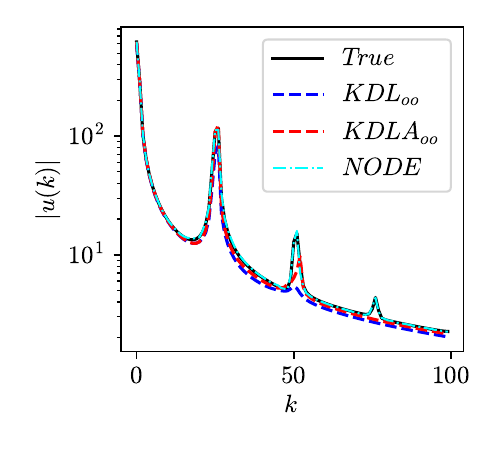}\\ 
  (c) & (d) & (e) \\
\end{tabular}
\end{center} 
\caption{\label{TW_beating_KS}
Predictions for  the Kuramoto-Sivashinsky with beating traveling waves with $L=29.30$. 
Panel (a) shows a snapshot of the dynamics predicted from the ground truth, \KDLoo, \KDLAoo, NODE, corresponding to top to bottom panels, respectively. Panel (b) compares the value of $u(x,t)$ from the different models  when  $u=u(-2.5,t)$, $u=u(0,t)$ and $u=u(2.5,t)$, corresponding from  top to bottom panels, respectively. Panels (c)-(e) represent the eigenvalues of the Koopman operator for \KDLAoo, the system energy, and the power spectrum of the model's predictions and true data, respectively.
} 
\end{figure}

The top panel of figure \ref{TW_beating_KS}a shows the complex structure of the data (true trajectory), in which a beating traveling wave is observed. The other three panels show the trajectory that results from evolving the same initial condition from the top panel forward on time using \KDLoo, \KDLAoo and NODE, from top to bottom, respectively. Overall, the three methods can capture qualitatively the dynamics.
Figure \ref{TW_beating_KS}b shows the values of the velocity at different spatial locations over time for different model's predictions and true values (e.g., $u(x=-2.5,t)$, $u(x=0,t)$ and $u(x=2.5,t)$). This figure displays the  decaying dynamics from the  the \KDLoo's predictions over a span of $250$ time units, as it is also observed in the temporal evolution  of the energy  (displayed in \ref{TW_beating_KS}d). \KDLAoo outperforms the traditional \KDLoo whose dynamics gradually decay as time evolves.
Figure \ref{TW_beating_KS}c shows the eigenvalues of the {Koopman operator} for \KDLAoo lie near the unit circle\ALrevise{, resulting in a much less evident decay than exhibited by \KDLoo}. Finally, figure \ref{TW_beating_KS}d shows the power spectra for the true data and the two models. 
The predictions of \KDLAoo are capable of capturing very accurately the structure of the data, such as the traveling and beating periods, and outperforms \KDLoo.
\ALrevise{In total,} figure \ref{TW_beating_KS} shows that both the state space and function space approach \ALrevise{(when using \KDLAoo)} lead to excellent agreement with the true data.

\subsubsection{\textcolor{black}{Kuramoto-Sivashinsky with chaotic dynamics }}

\begin{figure}
\begin{center} 
\begin{tabular}{c}
\includegraphics[width=0.6\linewidth]{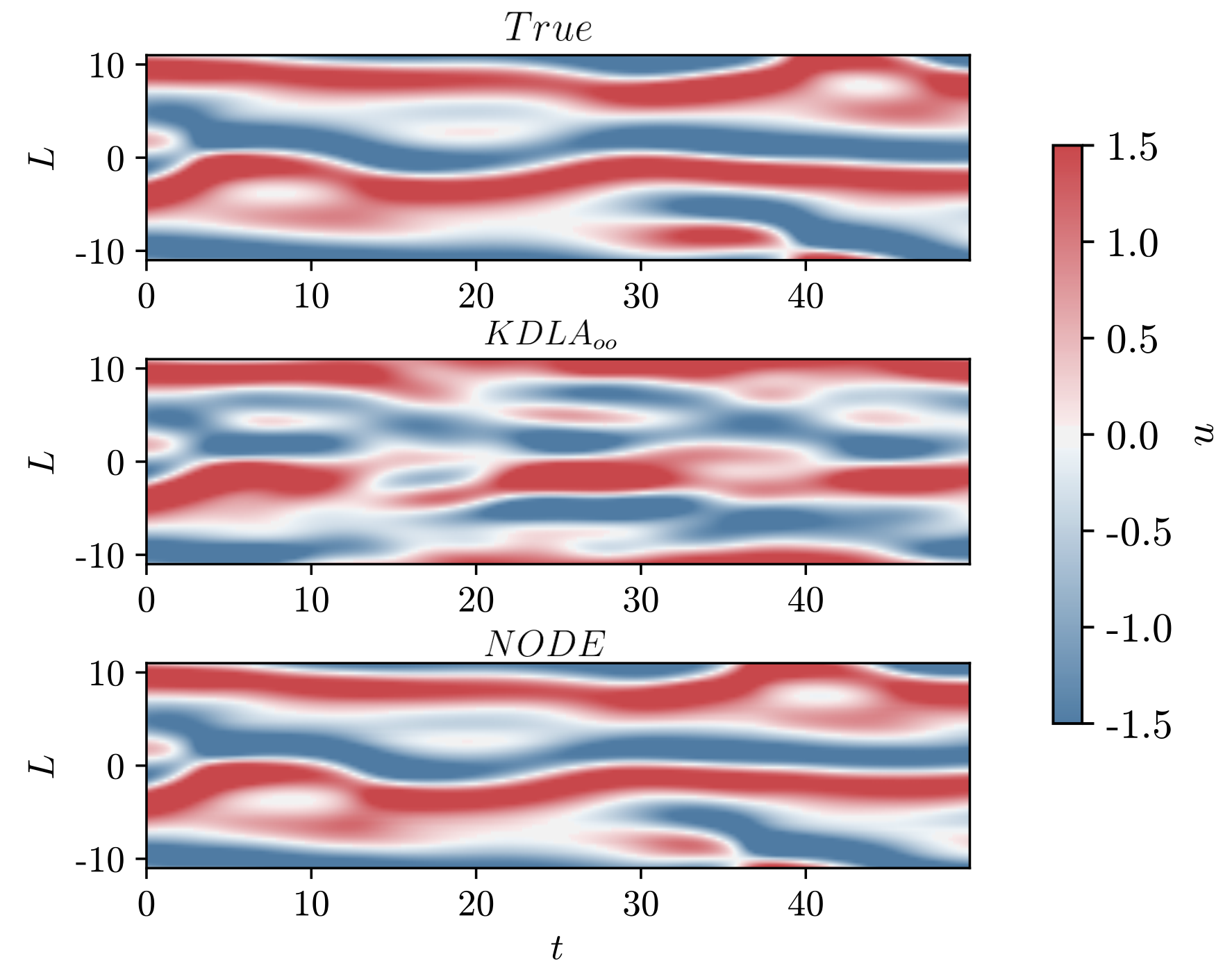}\\
(a)
\end{tabular}
\begin{tabular}{ccc}
\includegraphics[width=0.33\linewidth]{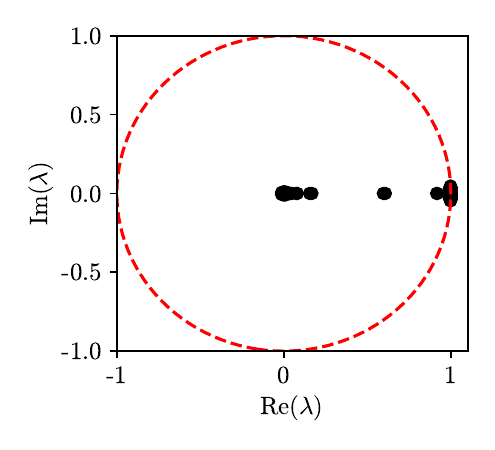} &
\includegraphics[width=0.33\linewidth]{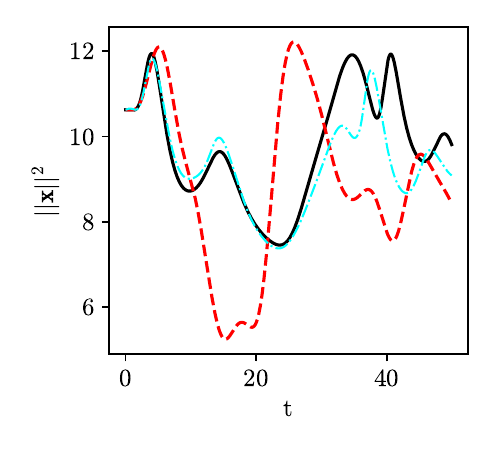}&
\includegraphics[width=0.33\linewidth]{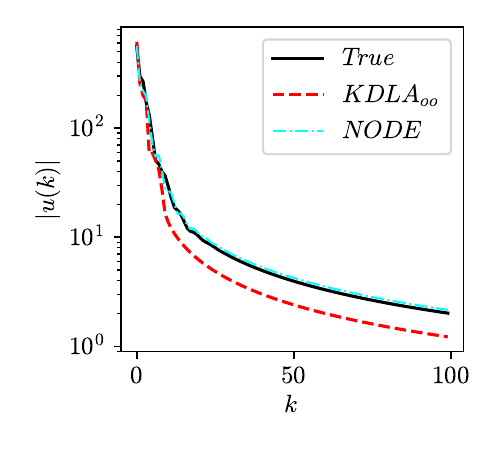}\\
(b) &(c) & (d)\\
\end{tabular}
\end{center} 
\caption{\label{KS_L22} \textcolor{black}{ 
Predictions for  the KSE with 
$L=22$.
Panel (a) shows a snapshot of the trajectory given by the ground truth, \KDLAoo and NODE, corresponding to top, middle and bottom panels, respectively. 
Panels (b)-(d) display the eigenvalues of the Koopman operator for the \KDLAoo, the system energy and the power spectrum of the model's predictions and true data, respectively. }}
\end{figure}

\textcolor{black}{
In our final case with the KSE, we \ALrevise{consider the}
chaotic regime of $L=22$.
We compute the trajectory with 64 Fourier modes and assemble a dataset from a single trajectory up to $t=10^3$ \ALrevise{time units}, separated \ALrevise{every} $\delta t=0.05$. We remove the transient dynamics from the dataset. We create  150 dictionary elements in addition to the state  $\textbf{x}$; thus, the set of observables $ \mathbf{\Psi }$ is 214. }

The top panel of figure \ref{KS_L22}a shows the \ALrevise{a trajectory from the KSE in this chaotic regime from a random initial condition evolved forward $50$ time units.}
The middle and bottom panels show the predicted trajectories when evolving the same initial condition with \KDLAoo and \ALrevise{the} NODE approaches, respectively.
\ALrevise{The \KDLAoo shows qualitative agreement for over 10 time units before diverging from the true solution. We can only ever expect short-time tracking when reconstructing chaotic systems because chaos causes small perturbations in trajectories to diverge. The timescale over which this divergence happens is known as the Lyapunov time, which is $\sim20$ time units in this case.}
Figure \ref{KS_L22}b shows the eigenvalues of the Koopman operator for \KDLAoo, where some of them are located on the unit circle, so the dynamics will take long times to decay.
\textcolor{black}{The new feature of this case is that the 
% infinite dimensional
Koopman operator has a continuous spectrum that the discretized (matrix) operator cannot completely capture, as demonstrated by the divergence of the trajectories at longer times}.

The NODE approach shows excellent agreement with the true data in the bottom panel of figure \ref{KS_L22}a, as well as the predicted system energy and power spectrum (see figure \ref{KS_L22}c-d, respectively). \textcolor{black}{ This agrees with the previous results from \citet{alec_chaos}, who showed that NODE works extremely well in terms of short-time tracking and long-time statistics for KSE with $L=22$.}
Finally, we remark that finding an accurate approximation of the Koopman operator for a chaotic system is still challenging due to its continuous spectrum, as we have demonstrated in this case. 
Recently, \citet{mpEDMD} presented a variation of the EDMD framework to account for the continuous spectrum in  dynamical systems.

\subsection{Stuart-Landau equation}

Finally, we consider a simple system with multiple equilibria.
We study the one-dimensional Stuart-Landau equation  which can be written as \cite{bagheri_2013,page_kerswell_2019}
\begin{equation}
 \dot{r}=\mu r-r^3
\end{equation}
which has a pitchfork bifurcation at $\mu=0$; for $\mu >0$, the system is characterised by an attractor at $r \pm \sqrt{\mu}$ and a repeller at $r=0$. Assuming $\mu >0$, the previous equation can be rescaled with $R=r/\sqrt{\mu}$ and $T=\mu t$ to

\begin{equation}
   \dot{R}=R-R^3
\end{equation}

\noindent
which has solution 
\begin{equation}
    R(T;R_0)=\frac{1}{\sqrt{1+b(R_0)e^{-2T}}}
\end{equation}
where $b(R_0)=(1-R_0^2)/R_0^2$, so $R(0;R_0)=R_0$.

Our training set consists of a single trajectory evolved up to $t=20$ with a $\delta t=0.04$ \ALrevise{resulting in 500 snapshots of data.}
We create 25 dictionary elements in addition to the state $\textbf{x}$; thus, the set of observables $ \mathbf{\Psi }$ is 26. 
\textcolor{black}{We remark that \citet{page_kerswell_2019} also considered this case with EDMD in which the observable vector was constructed from polynomials in $R$, $\Psi(R)=(R,R^2,R^3,R^4)$. In their study, they revealed a limitation: with this specific set of observables, EDMD failed to capture the critical crossover point, located at $R=1/\sqrt{2}$ between the repelling and attractor equilibria. However, \KDLAoo utilizes a more extensive dictionary, allowing us to overcome this limitation and successfully capture the crossover point.}

\begin{figure}
\begin{center} 
\begin{tabular}{cc}
\includegraphics[width=0.4\linewidth]{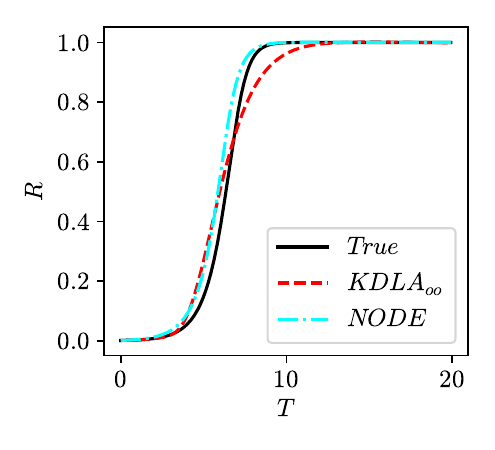}&
\includegraphics[width=0.4\linewidth]{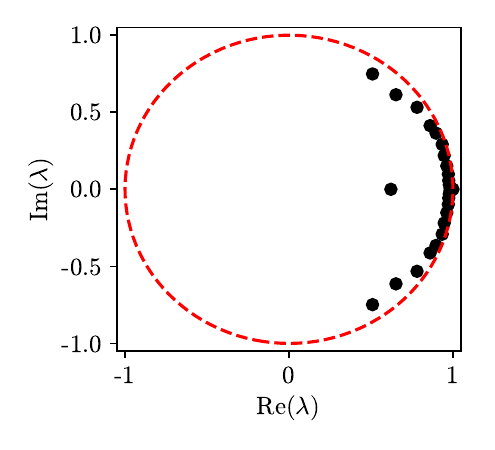}\\
(a) & (b) \\
\end{tabular}
\end{center} 
\caption{\label{SL_fig}
 Predictions for the Stuart-Landau equation with $R_0=10^{-3}$.
Panel (a) shows the trajectories for the true and predicted dynamics from the \KDLAoo and NODE.  Panel (b) represents the eigenvalues of the Koopman operator for the \KDLAoo. 
} 
\end{figure}

Figure \ref{SL_fig}a shows the performance of \KDLAoo and NODE by comparing the temporal evolution of a single trajectory. It shows the predictive capabilities of both models as they can capture the repelling and attractor equilibria accurately.
Figure \ref{SL_fig}b displays the eigenvalue spectrum of the Koopman operator. This example shows a promising approach to tackle similar behaviour observed along heteroclinic connections between equilibria of the Navier-Stokes equations.

\section{Conclusions}

\MDGrevise{Methods based on data-driven approximation of the Koopman operator for dynamical systems are promising for the prediction of the time evolution of systems with complex dynamics.
A widely used approach, extended dynamic mode decomposition with dictionary learning (EDMD-DL), is modified here to simultaneously determine the dictionary of observables and the corresponding \ALrevise{Koopman operator.}
The method leverages automatic differentiation to perform gradient descent through the Moore-Penrose pseudoinverse. Furthermore, we consider the performance of a ``pure" Koopman approach, which performs time-integration of the linear, high-dimensional system governing the evolution in the space of observables, as well as a modification where the system alternates between spaces of states and observables at each step. The latter case no longer represents linear evolution, somewhat diminishing its connection to the true Koopman operator representation, but can lead to substantial improvements in time evolution predictions over the purely linear case. In fact, the study that introduced EDMD-DL \cite{Qianxiao} described the former approach, but the code accompanying the article implemented the latter.}
We have compared these approaches to EDMD-DL, 
for 
\MDGrevise{systems ranging from two and three-dimensional systems of ODEs with steady, oscillatory and chaotic attractors to partial differential equations with increasingly complex behavior.}
We have enhanced our analysis by incorporating comparisons with a state space approach using the neural ODE approach. The NODE leads to a superior performance in predictions when the function state approach evolves exclusively within the space of observables. However, when the function state approach alternates between the spaces of state and observables, we have found that its performance is comparable to the state space approach.

The framework presented herein offers new avenues and challenges for data-driven dynamical systems.  The natural extension of our framework is to further enhance its accuracy and physical relevance by incorporating physical laws into our methodology such as symmetries or conservation of properties (energy). A way to incorporate those physical laws is by including relevant observables for the systems being studied. For example, in Hamiltonian systems, the conservation of total energy could be incorporated to improve the predictive power of the method. It is worth noting that the training data used in this study is free of noise, which may not be the case for real-world datasets. 
\textcolor{black}{By adopting a linear formulation in the Koopman analysis, then the lack of nonlinearities simplifies the mathematical representation of the system's evolution.
This simplification could potentially lead to more stable and predictable behavior, especially in the presence of noise, as the linear model may not amplify disturbances in the same way nonlinear models might.}
Incorporating noise into the training data could help make the method more robust to real-world problems. The primary goal of our future studies is to develop a Koopman-based approach for accurately predicting high-dimensional chaotic systems, such as the Kuramoto-Sivashinsky equation with a large number of modes. Ongoing research is focused on refining and optimizing the methodology to achieve this goal, including exploring the impact of different network architectures and optimization algorithms.

\section*{Appendix}

% \subsection*{Appendix A}

% \textcolor{black}{
% In this appendix, we present the singular values of the Koopman operator for \KDLAoo  applied to the Duffing oscillator.
% Figure \ref{SV}a shows the space-time tracking of the  singular spectrum of the Koopman operator matrix at various iterations during training. We observe that the singular values led to a full-rank matrix. In Figure \ref{SV}b, we present the absolute difference between consecutive singular values from figure \ref{SV}a for the the Koopman operator post-training. Those singular values are not close enough to lead to numerical problems while computing the gradients gradient through the pseudo inverse}

% \begin{figure}
% \begin{center} 
% \begin{tabular}{cc}
% \includegraphics[width=0.5\linewidth]{Revision/S.pdf}&
% \includegraphics[width=0.5\linewidth]{Revision/Diff.pdf}\\
% (a) & (b) \\
% \end{tabular}
% \end{center} 
% \caption{ \textcolor{blue}{
% (a) `Space-Time' tracking of the singular spectra of the Koopman operator at at various iterations during training, trained on the Duffing oscillator. (b) Absolute discrete difference for consecutive singular values for the Koopman operator post-training.  \label{SV}
% } }
% \end{figure}

\subsection*{Appendix A}

\begin{figure}
\begin{center} 
\begin{tabular}{cc}
\includegraphics[width=0.5\linewidth]{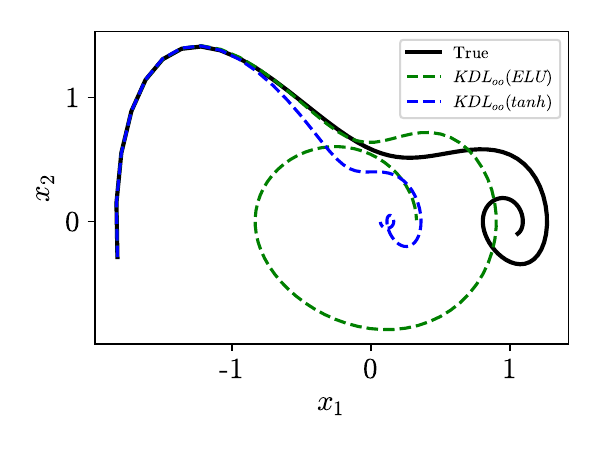}&
\includegraphics[width=0.5\linewidth]{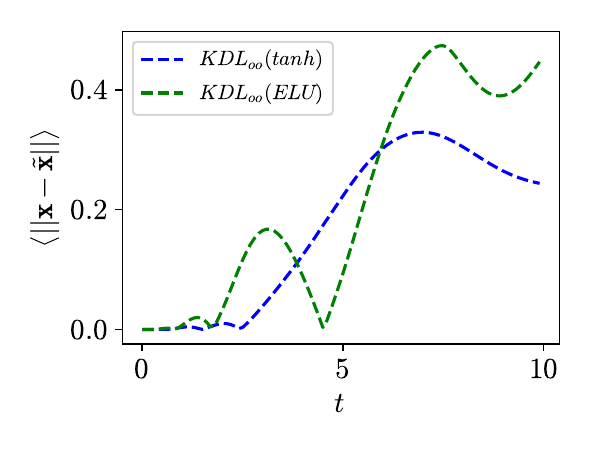}\\
(a) & (b) \\
\end{tabular}
\end{center} 
\caption{
\textcolor{black}{
Predictions for the Duffing oscillator with \KDLoo with the architecture from Table 1 (3 layers of 100 nodes, 100 dictionary elements, and ELU as activation function), and the configuration used in \citet{Qianxiao} (3 layers of 100 nodes, 25 dictionary elements, and tanh as activation function). Panel (a) shows the predicted trajectories from both architectures  and the true data. Panel (b) shows the short-time error computed by taking the mean squared difference of 1000 different initial conditions between the exact trajectory and the different models as a function of time. \label{appe_fig}}
} 
\end{figure}

\textcolor{black}{
In this appendix, we compare  two distinct  architectures for  \KDLoo~  applied to the Duffing oscillator.  Figure \ref{appe_fig} presents  the trajectory of the same  initial condition presented in figure \ref{duff}a  and the normalized ensemble-averaged tracking error for 1000 initial conditions. 
% Our goal is to provide a thorough and equitable evaluation of the performance of the two architectures within the context of  $KDL$.
The architecture used in \citet{Qianxiao} (3 layers of 100 nodes, 25 dictionary elements, and tanh as activation function) yields superior results compared the architecture used for $KDLA$  (3 layers of 100 nodes, 100 dictionary elements, and ELU as activation function). This discrepancy in performance is the primary rationale behind our decision to employ the former architecture when presenting the results in figure \ref{duff}.}\\

\section*{Acknowledgments}

This work was supported by ONR N00014-18-1-2865 (Vannevar Bush Faculty Fellowship).
\textcolor{black}{We acknowledge Prof Qianxiao Li for sharing his original EDMD-DL code \citep{Qianxiao}.}

\section*{Data Availability Statement}
The data that support the findings of this study are available at github: \url{https://github.com/rcrc15/KDLAoo}.
%from the corresponding author upon reasonable request.

%

\bibliographystyle{apalike} 
\bibliography{references.bib}
 
\end{document}